\documentclass[letterpaper, 10pt, conference]{ieeeconf}

\IEEEoverridecommandlockouts
\usepackage[dvips]{graphicx}
\input{epsf}

\usepackage{url}
\usepackage{subfig}
\usepackage{latexsym}
\usepackage{lastpage}
\usepackage{cite}
\usepackage{color}
\usepackage{listings}

\usepackage{epsfig}
\usepackage{amsfonts}
\usepackage{amsmath}
\usepackage{amssymb}
\usepackage{mathtools}
\usepackage{algorithm}
\usepackage{algorithmic}
\usepackage{latexsym}

\newcommand{\beq}{\begin{equation}}
\newcommand{\eeq}{\end{equation}}
\newcommand{\beqr}{\begin{equation}\begin{array}{l}}
\newcommand{\eeqr}{\end{array}\end{equation}}
\newcommand{\beqa}{\begin{eqnarray}}
\newcommand{\eeqa}{\end{eqnarray}}

\newcommand{\be}{\begin{equation}}
\newcommand{\ee}{\end{equation}}
\newcommand{\bea}{\begin{eqnarray}}
\newcommand{\eea}{\end{eqnarray}}
\newcommand{\ba}{\begin{array}}
	\newcommand{\ea}{\end{array}}
\newcommand{\beas}{\begin{eqnarray*}}
	\newcommand{\eeas}{\end{eqnarray*}}
\newcommand{\leftm}{\left[\begin{array}}
	\newcommand{\rightm}{\end{array}\right]}

\definecolor{gray}{rgb}{0.4,0.4,0.4}
\definecolor{darkblue}{rgb}{0.0,0.0,0.6}
\definecolor{cyan}{rgb}{0.0,0.6,0.6}
\definecolor{applegreen}{rgb}{0.55, 0.71, 0.0}
\lstdefinelanguage{XML}
{
	morestring=[b]",
	morestring=[s]{>}{<},
	morecomment=[s]{<?}{?>},
	stringstyle=\color{black},
	identifierstyle=\color{darkblue},
	keywordstyle=\color{cyan},
	morekeywords={xmlns,version,type}
}


\begin{document}
	\graphicspath{{./figures/}} 
	\title{An Embodied, Platform-invariant Architecture for Connecting High-level Spatial Commands to Platform Articulation}
	\author{A. Jang Sher, U. Huzaifa, J. Li, V. Jain, A. Zurawski, and A. LaViers \thanks{Email: alaviers@illinois.edu}\thanks{Five authors are with the Department of Mechanical Science and Engineering, University of Illinois at Urbana-Champaign, IL 61801, USA.  J. Li is with the Department of Electrical and Computer Engineering, University of Illinois at Urbana-Champaign, IL 61801, USA.}}
	\maketitle	
	
	\begin{abstract}
	In contexts such as teleoperation, robot reprogramming, human-robot-interaction, and neural prosthetics, conveying movement commands to a robotic platform is often a limiting factor. Currently, many applications rely on joint-angle-by-joint-angle prescriptions.  This inherently requires a large number of parameters to be specified by the user that scales with the  number of degrees of freedom on a platform, creating high bandwidth requirements for interfaces.  This paper presents an efficient representation of high-level, spatial commands that specifies many joint angles with relatively few parameters based on a spatial architecture that is judged favorably by human viewers.  In particular, a general method for labeling connected platform linkages, generating a databank of user-specified poses, and mapping between high-level spatial commands and specific platform static configurations are presented.  Thus, this architecture is ``platform-invariant'' where the same high-level, spatial command can be executed on any platform.  This has the advantage that our commands have meaning for human movers as well.  In order to achieve this, we draw inspiration from Laban/Bartenieff Movement Studies, an embodied taxonomy for movement description.  The architecture is demonstrated through implementation on 26 spatial directions for a Rethink Robotics Baxter, an Aldebaran NAO, and a KUKA youBot. User studies are conducted to validate the claims of the proposed framework.  
	\end{abstract}

	\section{Introduction}\label{introduction}
	  Leveraging human intelligence can overcome challenges posed by dynamic environments in contexts such as teleoperation in disaster settings, reprogramming robotic systems in factories, and neural robot interfaces embedded in human bodies.  Each of these scenarios requires communication of complex movement designs from a human to a machine.  Here, a bottleneck exists.  While humans can share complex movement instructions with each other -- as is done from choreographers to their dancers or someone giving instructions over the phone, methods to recreate this on capable robot hardware do not exist. 
	  
	  Command representation, which is typically created in a joint-angle-by-joint-angle command architecture, is often a limiting factor: either requiring labor-intensive command creation that limits the rate of commands or resulting in high demands on interface affordances. Some of the challenges faced in teleoperating robots can be seen as a twofold issue: 1) lack of \textit{expressivity} in command architectures and 2) disparity between humans' perception of movement and the commands communicated to the robots to generate movement -- a kinesthetic mismatch. 
	
	
	Teleoperation presents various challenges for the human operator with the added impediment of remote perception and manipulation \cite{chen2007human}. Research groups are tackling this difficulty by incorporating intuitive gaming controllers or datagloves \cite{hu2005robot, turner2000development, song1999share}. Even though controlling an articulated robot with limited sets of knobs keeps the task tractable for an operator, it limits the functional potential of a robot. 
	
	\begin{figure*}[ht]
\centering
\includegraphics[width=\textwidth]{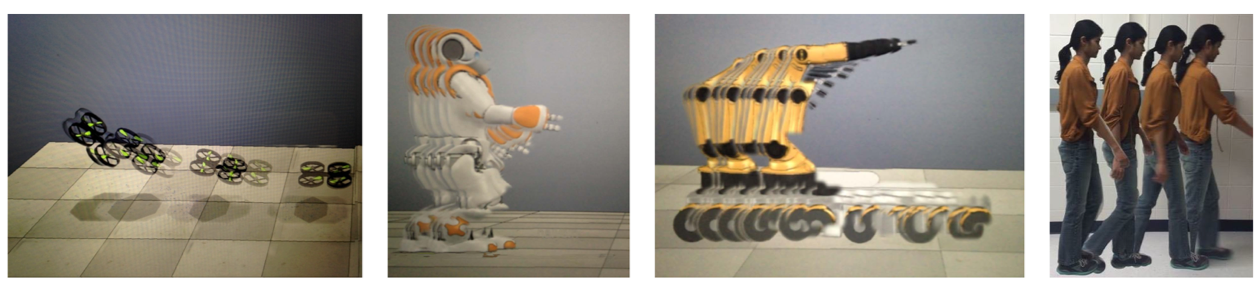}
\caption{Left to right: AR Drone quadcopter, Aldebaran NAO, KUKA youBot, and a human moving `forward.' The hand-generated images illustrate how spatial commands can elicit translation \textit{and articulation}.  Each platform is executing the same high-level idea but is using a different series of configuration changes to achieve it.} 
\label{motion_primitive}
\end{figure*}
	
For example, in the iRobot PackBot \cite{yamauchi2004packbot}, the control interface restricts simultaneous translation and arm articulation, limiting the range of motion at the disposal of the operator in order to map the command architecture to a hand held controller.  This platform serves as a particular point of inspiration for this work.  We note two common tasks faced by soldiers in the field that the current hardware \textit{is} capable of achieving but the human-machine interface \textit{does not} support:
\begin{itemize} \itemsep0pt
\item sweeping terrain to uncover buried wires or other feature; \textit{requires traversing ground with simultaneous arm articulation perpendicular to the path of travel}
\item removing the covering of suspected explosive devices, while getting the platform out of a potential blast radius; \textit{requires moving arm away from device while simultaneously driving away from the device}.
\end{itemize}
Note that both of these tasks \textit{require simultaneous articulation of the arm and translation of the base}. Currently, soldiers switch between interaction modes of a video game controller in order to control \textit{either} the base or the arm.  We will propose a framework that unifies articulation and translation commands under consistent spatial directions.
	
Given a desired end effector position and orientation for a robot, there are a number of ways to determine the joint angle for the robot, inverse kinematics (IK) being one of them. This is a well-studied problem with analytical as well as numerical solvers for the inverse kinematics problem \cite{welman1993inverse, girard1985computational, bodenheimer1997process}. However, there are inherent issues with IK if used on its own. For example, in multi-link manipulators, some end-effector poses introduce singularity in the system making it hard to find the exact corresponding joint angles \cite{corke2011robotics}. Further, this problem is under-determined so multiple joint parameter specifications can produce the same end effector position. 

Moreover, not all of the produced configurations are desirable for the given task. Allowing a human operator to select of the most desirable pose for the robot still presents challenges. In numerical optimization algorithms, a configuration space pose is found by iteratively minimizing an objective function.
However, this criteria might not align with what the human operator intended. Work in \cite{grochow2004style} aims at producing poses using stylized IK where the system gives preference to poses `similar' to poses in the extensive training data set. 

For robots with multiple serial chains, there are techniques rooted in IK-based formulations of manipulators \cite{lee2014relative}. In this case, a relative Jacobian defines the rate of change of distance between two end effectors. However, issues similar to IK are faced here as well. Furthermore, in the case of humanoids or robots with more degrees of freedom, solving for IK becomes computationally more expensive \cite{TolaniGB00}.\par
There has also been extensive work in imitating human motion on robots -- especially humanoid robots. Some prominent work in real-time motion generation can be found in \cite{kimyouiros09}, \cite{freiburg_imit_icra14}, and \cite{Choi2009}. These mostly involve kinematic mapping and suitable scaling of human joints and links, respectively, to their most likely counterparts in a given humanoid robot. Some related work in generating movement from human data can be found in \cite{kimparkiros06}. \par

	Poses for a platform are often generated using model-based approaches. Using model-driven approaches, ideas have been explored for parameterizing robot platforms so that it may be easier to generically analyze them. In this line of work, classical Denavit-Hartenberg parameters for robot arms \cite{DH} and Unified Robot Description Format (URDF) \cite{urdf} for any robot are established methods that span many or all, respectively, specific platforms. On the other hand, data-driven approaches have shown tremendous promise in applications such as grasping \cite{goldfeder2011data,bohg2014data}. 
	
	A promising approach to make the robots perform complex movements is through composition of simple \textit{motion primitives}. For this purpose, we need special tools for modeling and control that abstract away low-level details of platform control. For example, specifications phrased in linear temporal logic (LTL) have sequenced robotic motions for a given high-level task such as surveillance \cite{ozay2011distributed}, urban driving \cite{chen2013formal}, and stylistic movement \cite{laviers2011IEEERAM}. This work relies on abstraction, following the over-arching aim of \textit{symbolic} control \cite{belta2007symbolic}. For approaches to create these objects from samples of discrete behavior, work has been done to build automata with user queries \cite{biermann1972synthesis,bollig2010libalf}, leading to data-driven approaches \cite{gillies2009learning}. User-inspired {motion primitives} are explored in \cite{Delvecchio03class,drumwright2004exemplar,nakaoka2004leg,hofmann2006robust,guerra2007language,matsubara2010learning,cohen2011planning}. 
These methods for movement generation in robots have to be tailored for the platform at hand. Therefore, a forward movement command, for example, is customized to a given platform architecture.

In parallel, movement notation systems such as Labanotation have been used to generate motion programs for robots \cite{abe2017use}.  This work is inline with human motion imitation.  Labanotation is a system that stemmed from the work of Rudolph Laban \cite{guest2014labanotation} and is a detailed recording of a specific sequence of human motion.  Typically it is used for archival records of dances, which contain detailed instructions and intentions of a choreographer that may not even render well in video records.   In \cite{salaris2017robot} Labanotation is used to generate a specification language for the humanoid robot Romeo.  More detailed comparison to this work and existing methods will be made in Section \ref{comparison}.  Other movement notation schemes, such as Benesh Notation, have been proposed as motion specification for robots as well \cite{laumond2016dance}.

	 In this paper, we propose a description of movement that is quantitative, yet independent of a particular set of actuators, in order to formulate a high-level motion primitive.  In our framework, a primitive is defined by a desired final pose that will be achieved through low-level controllers on board a robotic platform.  Each primitive is endowed with properties of movement that can be applied to multiple platforms. In other words, a motion primitive should be thought of as a ``movement idea'' that is  \textit{platform-invariant} and only approximated by the platform performing it (see Fig. \ref{motion_primitive}). 
		
	Where should these movement ideas come from?  We choose the comprehensive movement system of Laban/Bartenieff Movement Studies (LBMS) \cite{Newlove2004laban}, \cite{studd2013}.  This system was also initiated by the work of Rudolph Laban but utilizes distinct training programs from those of Labanotation.  The work of LBMS is focused on embodied participation and experience of movement, whereas Labanotation emphasizes the theory of a complex set of symbols. 
	
	Our claim is that the LBMS system provides a way to encapsulate an \textit{embodied}\footnote{This means that the perspective requires a person to move in their own body to learn, use, or understand the perspective.  On the other hand, ``disembodied'' perspectives approach movement from an external point of view that ignores the role of the human body in order to learn, use, and/or understand the perspective.}  perspective of movement. LBMS provides a comprehensive framework, from a kinesthetic\footnote{Similar, to ``embodied'', this modifier references the human kinesthetic sense, enabled by proprioception, which is, roughly, the sense and experience of the body in space.}  perspective, for movement analysis and notation with four broad categories: Body, Shape, Space and Effort.  For the purpose of this paper, we will focus on the Body and Space categories to guide our approach to how humans organize specifying the \textit{what} and \textit{where} of motion.
	
	Various robotics research groups have leveraged the teachings of LBMS in their work; however, most prior work is focused on conveying the ``emotional'' state of a robot  \cite{nakata2002analysis, lourens2010communicating, knight2016laban,burton2016laban}. Here, our framework takes inspiration from LBMS to develop parameters that generate a wide array of poses.
		
	 Our approach will develop an extensive, flexible system with intuitive analogs for human movers to platform behavior.  Here, we focus on the final end configuration of a movement and intermediate steps to be defined by a basic, linear interpolation. ``Super-users'' will customize the command mapping through a choreographic installation process and using these commands novice ``users'' will generate a desired final configuration pose of a platform. Furthermore, our framework gives the super-user freedom to formulate movement ideas for their particular task. 
	   
The paper is organized as follows: Section \ref{lbms} describes the embodied taxonomy that informs the approach taken here.  Sections \ref{body} and \ref{space} detail the quantitative architecture we have developed. Next, we demonstrate the effectiveness of our framework in Section \ref{implementation} by implementing it on three different platforms: Rethink Robotics Baxter, Aldebaran's humanoid robot NAO, and KUKA's mobile manipulator youBot. Sections \ref{userstudy} and \ref{userstudy2} discuss two user studies conducted to validate our framework. Section \ref{comparison} gives more detailed comparison to existing methods.  Finally, Section \ref{conclusion} gives concluding remarks and outlines future goals.

\section{An Embodied Taxonomy for Movement}\label{lbms}

	This section will introduce concepts from Laban/Bartenieff Movement Studies (LBMS).  This taxonomy, developed by a community of dancers, choreographers, and physical therapists, has provided a key starting point for the work in the following sections.  Firstly, it bolsters the proposed notion that high-level commands can be platform-invariant.  Namely, the users of LBMS are human beings -- each a unique platform with differences in body part's length, muscular structures and development, and so on.  Yet, particularly in dance, there is a notion of `moving in unison' that requires that this taxonomy not prescribe each joint angle value, but instead higher-level movement `ideas' which can be translated to individual `movers' and yet \textit{be the same}.

		\begin{figure}[h!]
	\centering
	\subfloat[][]{\includegraphics[width=1.\columnwidth]{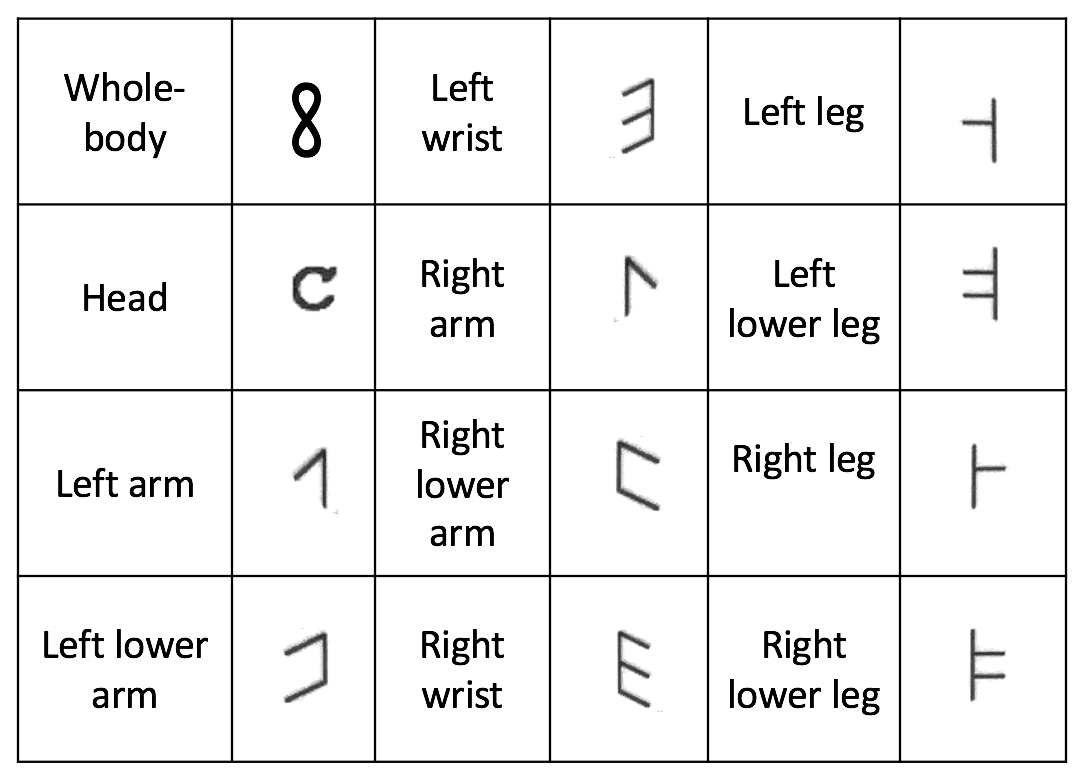}\label{bodysym}}
	\qquad
	\subfloat[][]{\includegraphics[width=1.\columnwidth]{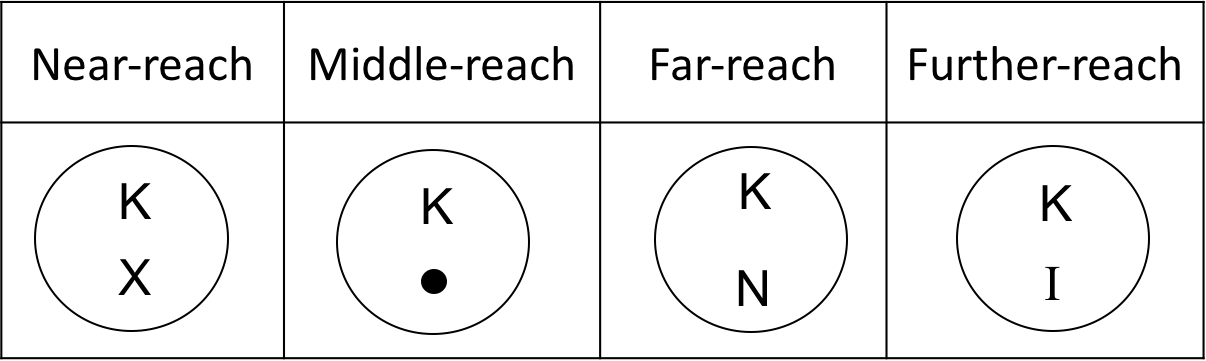}\label{kines_table}}
	\qquad	
	\subfloat[][]{\includegraphics[width=1.\columnwidth]{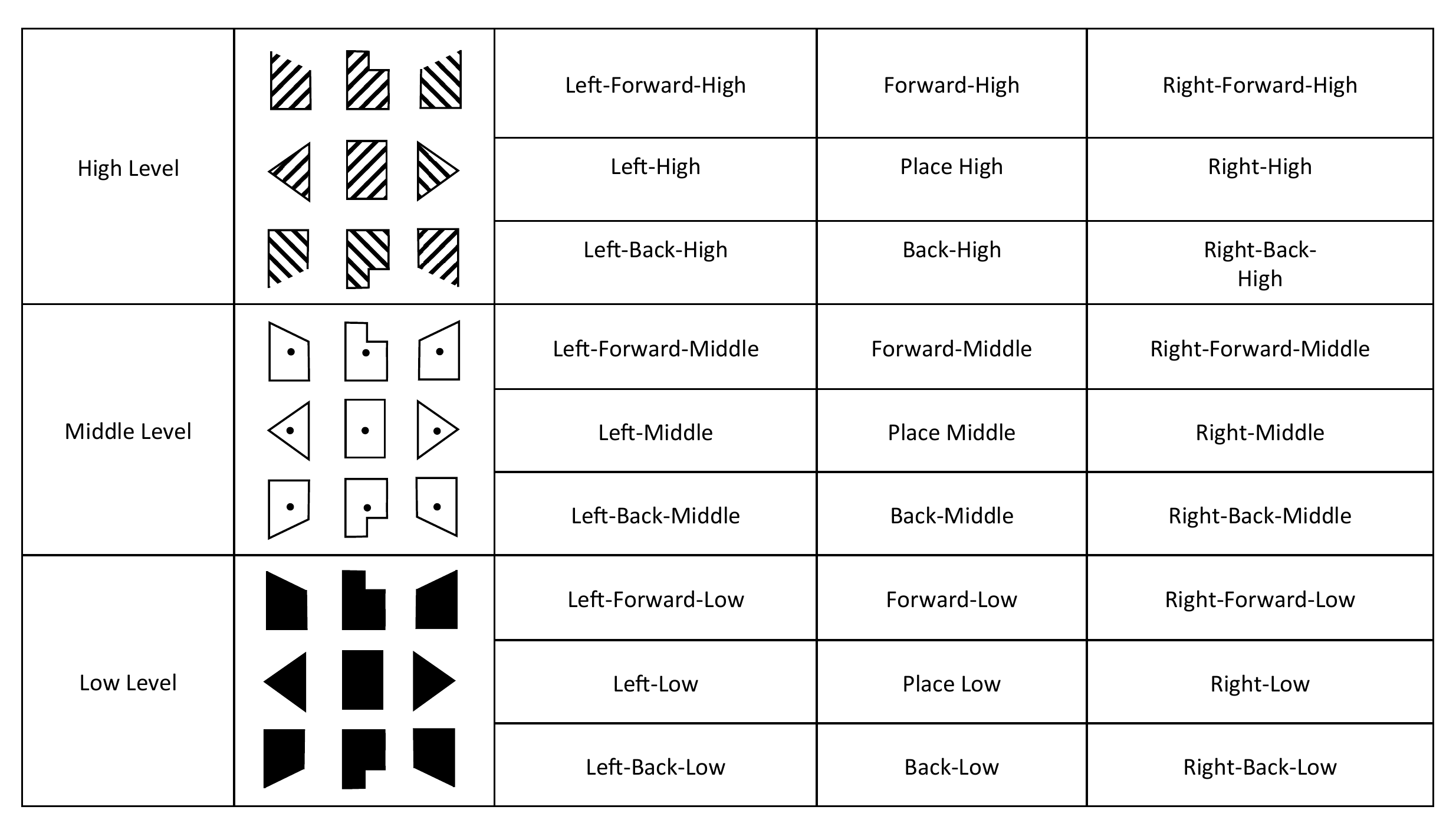}\label{direction_sym}}
	\caption{(a) This figure shows the Motif symbols of body parts that will be used here. (b) This figure shows symbols for varying kinesphere size. (c) This figure shows the 27 direction symbols from LBMS used in Motif and Labanotation, organized according to three zones, which correspond to the shading convention.  Each is inherently relative to a reference frame on the human body.  The body is relaxed or `neutral' in Place Middle.}
	\vspace{-.2in}
	\label{symbols}
	\end{figure}

The Body\footnote{Here, we will capitalize terms used from LBMS so as to distinguish them from their colloquial usages.} category describes the physical part of a mover participating in a motion. It covers 1) an anatomical naming scheme and terminology for each part and 2) combined behaviors of different parts of the mover.  In general, Body describes \textit{what changes for the body of a mover to carry out a movement}.  {Symbols that identify body parts are shown in Fig. \ref{bodysym}.}
In our work, this category inspires a flexible template for robot morphology. In Section \ref{body}, a generic labeling scheme is introduced that encompasses a large number of possible robot morphologies. This labeling scheme will allow us to organize recorded robot poses.  

%
%
%
	\begin{figure*}[ht]
 \centering 
 \begin{minipage}[t]{.4\textwidth}
 \centering
 \subfloat[]{\includegraphics[width=7.5cm]{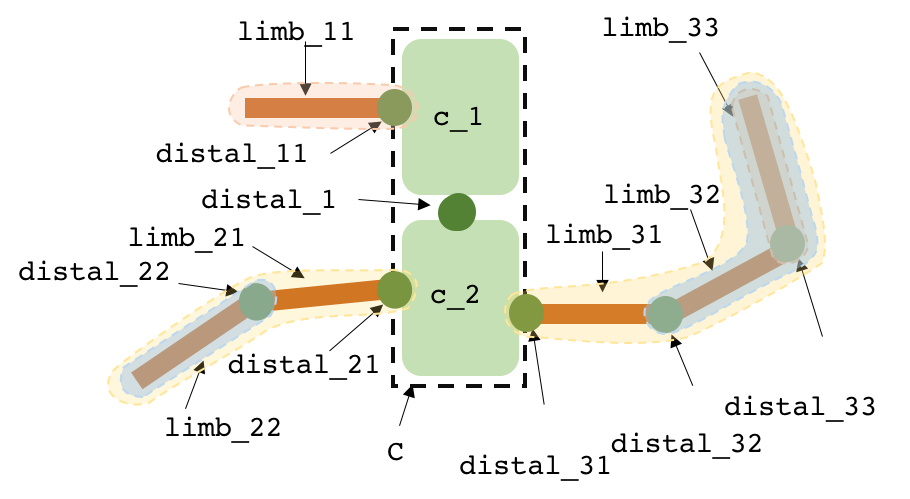}}
 \end{minipage}
 \begin{minipage}[t]{.28\textwidth}
 \centering
 \subfloat[]{\includegraphics[width=5.6cm,height = 5cm]{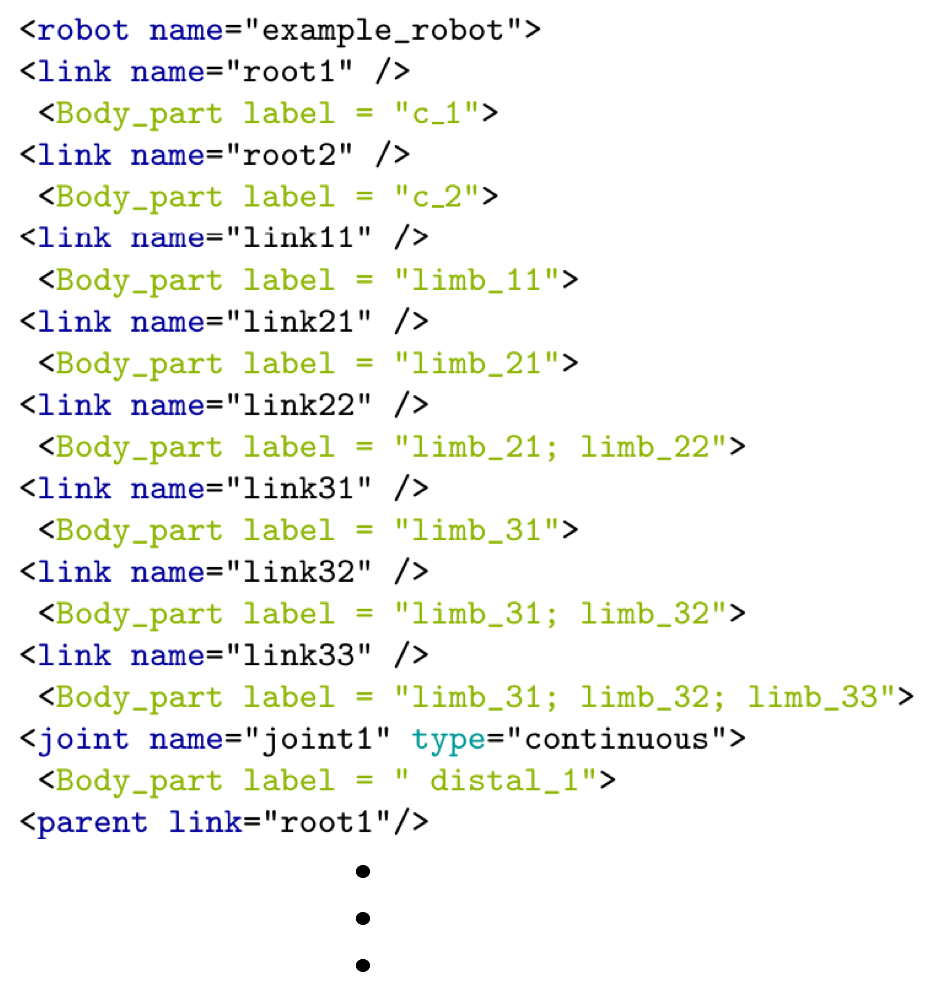}}
 \end{minipage}
 \begin{minipage}[t]{.28\textwidth}
 \centering
 \subfloat[]{\includegraphics[width=4.9cm, height = 5cm]{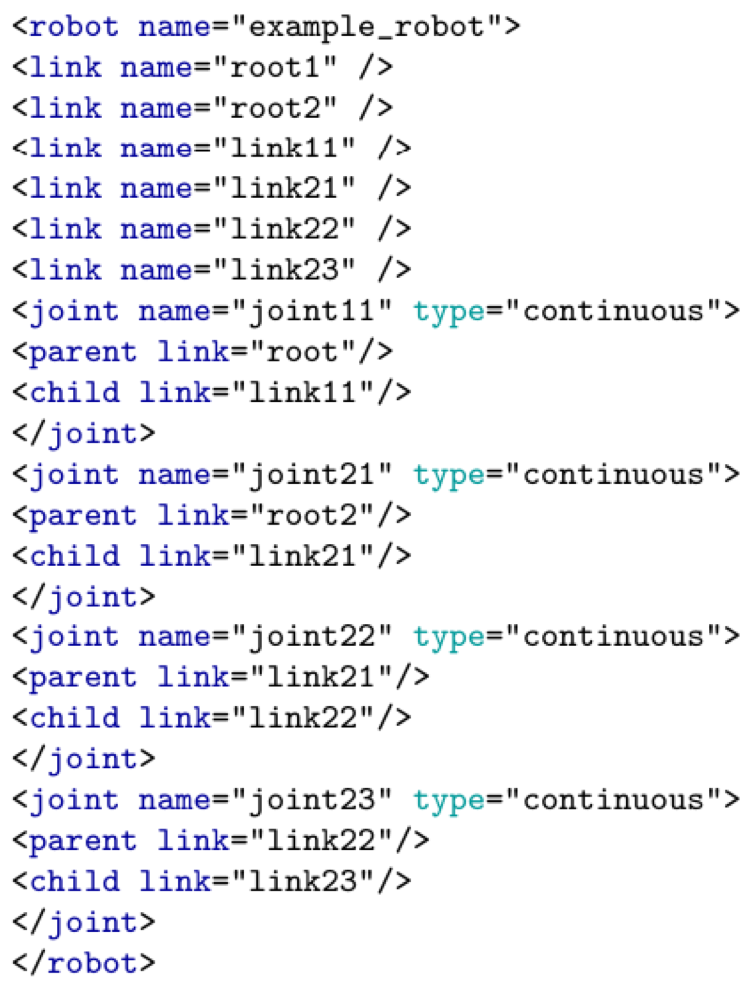}}
 \end{minipage}%

 \caption{A hierarchical, redundant platform labeling is demonstrated: (a) an asymmetric platform morphology with multiple parts in core and multiple links connected to the core; labels for $C$, $J$, and $L$ are shown; (b) a partial sample of the corresponding eURDF code where there are three sets of labels in \texttt{Body\_part} category: core, link(s) and joint labels; and (c) the corresponding URDF where individual link and joint location gets assigned a unique label.}
 
 \label{labels}
 \vspace{-0.15 in}
 \end{figure*}

The Space category of LBMS describes the spatial orientation -- or motivation -- of a motion.   Rudolph Laban 
had an almost mystic interpretation of the relationship between a mover and spatial points in their environment which inspired him to enumerate \textit{movement scales} as part of a larger understanding of balance in movement he termed Space Harmony.  These scales are analogous to musical scales used by piano players in warm up and training: they span a space of complex, related movements.  By practicing these scales, such as the Dimensional Scale, the A-Scale, the B-Scale, and the Diagonal Scale (defined in \cite{laban1966choreutics}), Laban felt that movers could find harmony in movement by training to more thoroughly span the points in ``the space that can be reached by easily extended limbs'' \cite{laban1966choreutics} or their \textit{Kinespheres}.  The system focuses on three types of reach space humans use: near-reach, middle-reach, far-reach, and further-reach (requiring locomotion to access); symbols for these concepts are shown in Fig. \ref{kines_table}. Further, it identifies the use of 26 specific pulls (as well a a neutral origin) which are listed in Fig. \ref{direction_sym}.
We are going to use this idea in creating a quantitative robotic framework: robots will be ``installed'' into our system when a user manipulates them through a similar process.


	Thus, the Body and Space categories are inherently interlinked. The body part involved in the movement is described by the Body category and the Space category specifies the proximity and the spatial direction in which a motion takes place.  More importantly, \textit\textbf{{this system explicates the fact that moving toward a spatial pull requires a complex body articulation.}}  In LBMS, this is practiced by movers via spatial scales that help individuals refine their so called \textit{access to space}. 
This highlights an important concept utilized in our work which is that access to Space is a learned, even trained, function achieved through Body articulation.  In LBMS, this is seen through the practice of movement scales, which outline a greater variety of Body-to-Space relationships than may be familiar or habitually visited by a mover.  In ballet, this is also seen through the practice of precise body positioning relative to the stage space \cite{warren}.  
Our work explicates this idea for artificial agents.
	
\section{Platform Description with a Flexible Labeling Scheme}\label{body}	
	In this section, we outline a method to describe robotic platforms in a redundant, hierarchical manner that will be leveraged in the next sections.  In particular, this labeling organizes robotic platform components into a few key categories that are needed to align with human labeling of links: the arm contains the forearm, for example.  Individual body components won't have a single label but may be given multiple labels depending on the position in the hierarchy. 
	
	First, note our terminology for the configuration space of the platform as $P$.  This discrete state space contains all kinematically possible robot configurations, i.e., it is the product space of all possible degrees of freedom as discretized by the actuator precision of each degree of freedom.  Thus, for a platform with $M$ degrees of freedom, if $p_{imin}$ and $p_{imax}$ are the joint angle limits with resolution approximated as $inc_i$, then $P=\{p_{1min},p_{1min}+inc_1,\hdots,p_{1max}\} \times ... \times \{p_{Mmin}, p_{Mmin}+inc_M,\hdots,p_{Mmax}\}$.
	
	In addition, we will enumerate three sets to group labels for the physical parts of a platform that will execute motion primitives: $C$, $L$, and $J$, where $C$ is a set of labels for the collection of parts that forms the platform's \textit{central} element(s); $L$ is a set of labels for \textit{linkages} that create a distinct platform shape which expresses itself in the environment; and $J$ is a corresponding set of labels for the \textit{joints} that articulate the groupings of links. 

	The elements in $C$ are the most centrally located pieces of the platform that must translate through space if the platform is to change locations with respect to a global, inertial frame.  This is the platform's `core' and corresponds closely with the location of the center of mass of the platform (as is the case in the colloquial notion of core in the human body).  
	On most platforms, this point is located within a rigid body. For platform labeling, we propose for the set $C$ to contain a label for that physical linkage containing the center of mass. In some robots, this core will thus contain moving linkages (i.e., wheels) as well. 
	
	Next, a platform may have linkages that extend from the platform.  Labels for these other links in the robot platform, $L$, are named $limb\_i$ where $i$ will indicate, using $core \in C$ as a root, the index of the limb in a hierarchical tree.  In particular, a label needs to be produced that applies to each possible subtree of the platform creating labeling overlap for individual links.  This limitation is similar to that of the method in \cite{luh1980line}.
	The set $J$ contains corresponding labels that describes each joint location consisting of a single or multiple joints.  Each joint is labeled as $distal\_i$ where $i$ matches the index of the associated limb, leaving a label for final contact point or terminal end effector. \par

	Using the above labeling, we are interested in a scheme that provides expandable and overlapping labeling of body parts. To this end, we have modified the widely used description format URDF. In this format, individual parts of a robot are described in an xml-like format with properties such as parents and children links for a particular robot part.  Our scheme appends to this system assigning broader, overlapping labels.\par
	The modified URDF includes a \texttt{Body\_part} label for each robot link and joint and will be referred to as Expressive Unified Robot Description Format (eURDF) from here onwards.
	In our eURDF mapping scheme, we need following information for each platform:
	\begin{itemize}
		\item Robot model description indicating joints, links, their limits and their locations, (i.e. information in the classical URDF, which implies $P$). 
		\item Overlapping label sets $C$, $J$, and $L$ for generating high-level commands.
	\end{itemize}
	A comparison of an eURDF  and a URDF file is shown in Fig. \ref{labels} corresponding to an example platform morphology. The URDF contains unique labels for individual links, cores, and joints. On the other hand, eURDF provides an overlapping, flexible labeling for the given platform. The platform under consideration has two parts $c\_1$ and $c\_2$ connected via joint $distal\_1$ signifying the core of the platform as shown by the dotted line around them. There are two sets of links connected to the core. All subtrees of links are assigned a label resulting in individual links containing multiple labels.  For this platform, the parameters $(C, J, L, P)$ are:
\vspace{-0.15in}
\begin{flalign*}
C &= \{ c\_1, c\_2 \} &\\
J &= \{ distal\_1, distal\_11, distal\_21, distal\_22, distal\_23 \} &\\
L &= \{ limb\_11, limb\_21, limb\_22, limb\_23 \} &\\
P &=\{p_{1min}, \hdots, p_{1max} \} \times \{ p_{2min}, \hdots, p_{2max} \}\\ &\qquad  \times \hdots  \times \{ p_{5min}, \hdots, p_{5max} \}
\end{flalign*}

Consider the Khepera IV robot that contains only two degrees of freedom, being two wheels housed in its base. To achieve a labeling useful for high-level, platform-invariant commands, the label may be assigned to the whole extent of the robot as core in set $C$ with empty sets $J$ and $L$, reflecting the fact that this platform cannot articulate in its environment. On the other hand, human movers, which have multiple links in the core and several layers of subtrees available in their limbs, will have large sets of labels for $C$, $L$, and $J$. Furthermore, the flexibility of our framework lies in not assigning a fixed set of labels for body parts of a platform. Different labels can be assigned for the same platform as deemed suitable for a task at hand.

\section{Data-driven Mapping Between High-level Pose Commands and Platform Configuration}\label{space}
		
With the overlapping labeling scheme outlined in Section \ref{body}, we now describe a method to generate a set of high-level, spatial commands that will be meaningful for any platform equipped with this labeling.  {As outlined for a soldier using a PackBot teleoperated robot in Section \ref{introduction}} and as seen in the practice of movement scales in LBMS described in Section \ref{lbms}, commands in space can be meaningful both for core translation and body articulation. 

This section will first define a generic model for Space-Body relationships for robotic platforms, termed a \textit{virtual space-access model (VSAM)}.  Second the section will propose the use of a relational database, an \textit{embodied configuration library (ECL)}, to store a mapping between these spatial commands and desired platform pose. The database will be customizable for different tasks and platforms by super-users and will enable high-level spatial sampling of the platform pose space $P$ by users with relatively few parameters. 	  

	\subsection{Virtual Space-access Model (VSAM)}\label{vsam}

\begin{figure}[!h]
		\centering
		\includegraphics[width=\columnwidth]{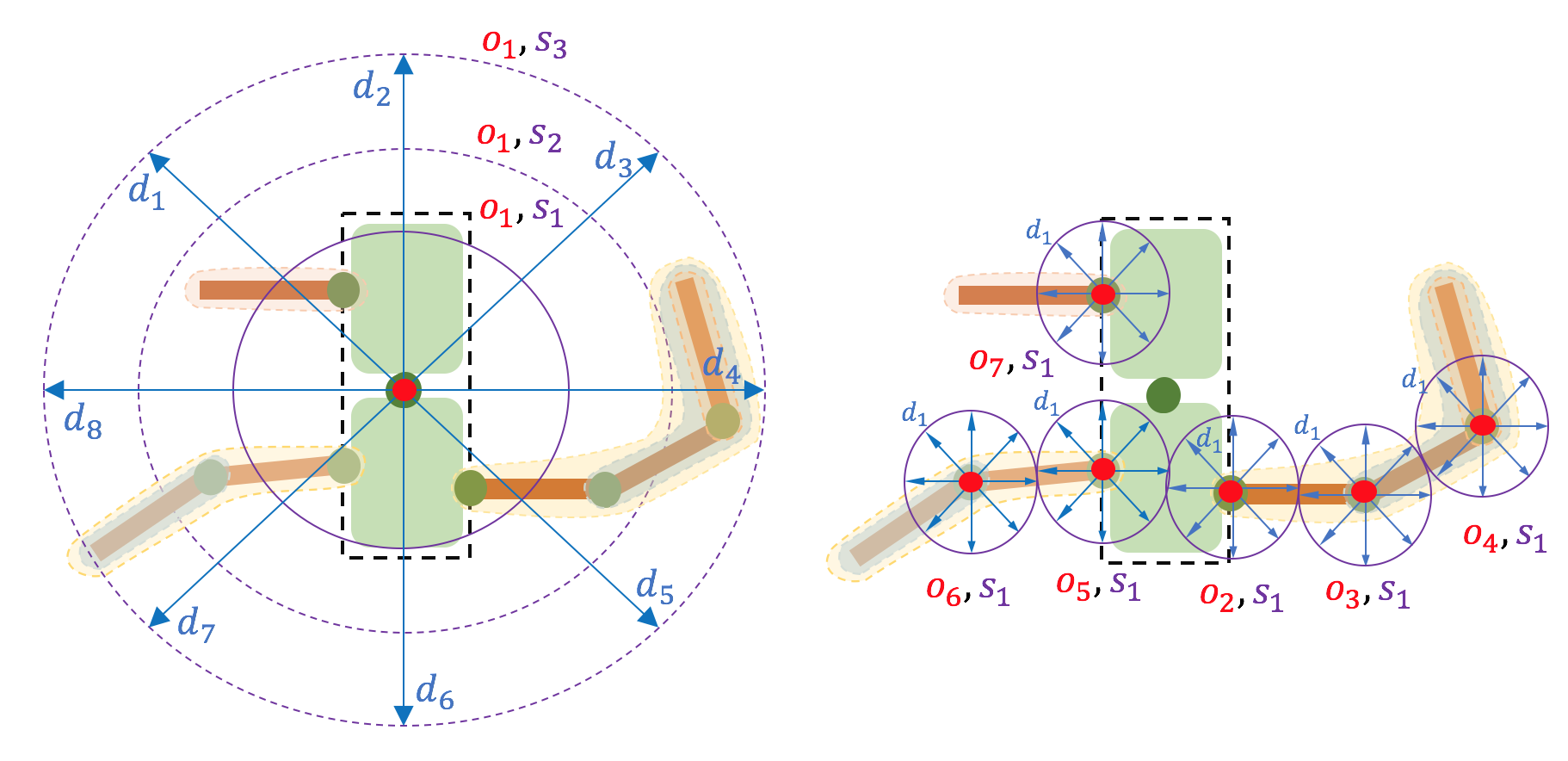}
		\caption{This figure illustrates an instantiation of a VSAM for the platform in Fig \ref{labels}, which is pictured here twice for clarity.  In {purple} are varying sizes $s_1$, $s_2$, and $s_3$; in red are origin locations $o_1, \hdots, o_7$; and in blue are directions $d_1, \hdots, d_8$. These parameters will determine a structure, and even meaning, for Space around the platform, delineating an artificial kinesphere for the platform.}
		\label{vsm}
	\end{figure}

We define virtual structure inside local geometric spaces within reach of the platform with a virtual space-access model.  First, a set of origins, which are located at joints on the platform (elements in $J$), where movement might be specified from, are defined in the set $k_o$.  Then, a set of directions, originating at any of these origins, are defined in the set $k_d$.  Finally, a set of sizes (or ``magnitudes'' of each directional vector) are established in $k_s$.  Thus, the triplet $(k_o,k_d,k_s)$ defines a VSAM. This VSAM can be applied to any platform.  Figure \ref{vsm} shows an example of a VSAM: it delineates varying sizes, established directions, and origin locations for the platform shown in Fig. \ref{labels}.   

Relative to an origin, $o\in k_o$, a limb $l\in L$ can move in any spatial direction $d \in k_d$ to generate a pose.   A spatial direction example could be the $z-axis$ of a body reference frame.  Each VSAM is indexed by $s\in k_s$ starting at $1$; each discrete pose thereafter will receive an ID incremented by one with the final pose in the library defining the farthest reach the platform can articulate for a particular direction, $(o, d, k_{s_{max}})$. Thereafter, an integer $x$ can be used to cause translation of the platform.  At $k_{s_{max}}+x$ the platform must translate according to an amount $x$ in the direction $d$ via a locomotion mode (if any available) in order to fulfill the command.  

	 The input then to the VSAM is a set of parameters $(l,o,d,s)$, which needs to output a unique joint configuration of the platform referred to as pose, $p \in P$. We'll call the space of such possible parameters the space of ``user parameters'' $UP$.  Then, a function maps the user parameters to a unique pose; that is $f_{SB}: UP \rightarrow P$.  The next section will define this mapping through a relational database.
\vspace{-0.2 in}

	\beqa
		\vspace{-0.2 in}
	p =f_{SB}(l,o,d,s) = \begin{bmatrix}
	\vspace{-0.05 in}
		p_1 \\[0.3em]
		\vspace{-0.05 in}
		p_2 \\[0.3em]
		\vspace{-0.05 in}
		\vdots  \\[0.3em]
		p_M
	\end{bmatrix}
	\label{framework}
	\eeqa

\subsection{Embodied Configuration Library (ECL)}\label{lib}

	We now take a data-driven approach to generate a large database of poses, which, through a relational schema, will define the mapping function, $f_{SB}$. This library is created corresponding to platform parameters $(C,J,L,P)$ and VSAM parameters $(k_o,k_d,k_s)$ (established by a super-user). By querying the database, a specific pose, $(p_1,...,p_M)^T\in P$, required to define the motion primitive is retrieved via user parameters $(l,o,d,s)$. Parameters $l$ and $o$ are labels from the eURDF labeling scheme (a link and joint, respectively); parameters $d$ and $s$ are integers.
	
A user can quickly access the stored mapping to poses (which may have tens of degrees of freedom specified -- and more for complex robotic platforms) via these four parameters, $(l,o,d,s)$.  This may result in a significant decrease in complexity to create robot configurations on the fly.  Super-users move the platform through a movement `scale' to set up the VSAM (or store poses in the database).  Motion is then generated by interpolation of the current pose to the user-specified final pose.  This can be achieved using existing path planning methods \cite{hwang1992potential, lavalle1998rapidly} and commercially available on-board controllers.

	\begin{figure}
		\centering
		\includegraphics[width=1.\columnwidth]{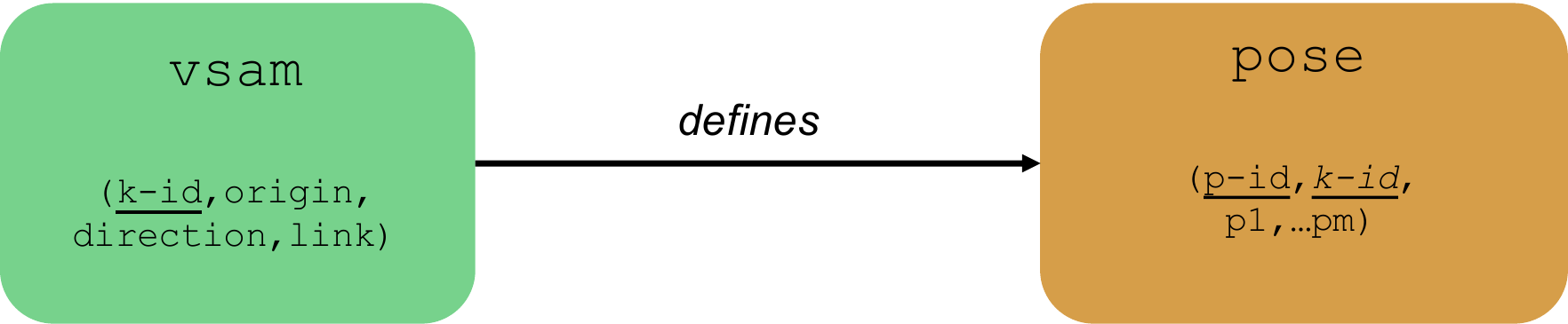}
		\caption{This figure shows an entity-relation (ER) diagram for the ECL. The databank provides a computational definition for $f_{SB}(\cdot)$, mapping user parameters to pre-stored platform configurations.}
		\label{library}
	\end{figure}
	
	The structure and size of the ECL of a platform is defined by the platform $(C,J,L,P)$ and the VSAM $(k_o,k_d,k_s)$, taking the form of a database of poses in terms of joint positions (in one table, \texttt{pose}) for a platform corresponding to user parameters, $(l,o,d,s)$ (in another table, \texttt{vsam}).  Figure \ref{library} shows this database schema.  
	This database relates joint positions to spatial directions, $d\in k_d$, for different body parts, $l \in L$, with origins, $o\in k_o$, indexed by size, $s\in k_s$. 
	These poses can be generated by manually moving each body part and automatically recording the poses or via some other automated or more precise method (discussed in Section \ref{conclusion}). The library is generated following the sequence of steps outlined in Algorithm \ref{alg:main}.
\setlength{\textfloatsep}{0pt}	
\begin{algorithm} 
\caption{Process for creating a spatial databank $ECL(C,L,J,P,k_o,k_d,k_s)$ for platform $(C,L,J,P)$}
\textbf{INPUT:} Off-line: super-user parameters $(k_o,k_d,k_s)$; on-line: user-parameters $(o,l,d,s)$\\
\textbf{OUTPUT:} Off-line: $f_{SB}: UP \rightarrow P$; online: $(p_1,...,p_M)^T$

\begin{algorithmic}[1]\label{alg:main}
\REQUIRE
\FOR{each pair $o\in k_o$ and $l\in L$}
	\STATE Select the subtree defined by $l$ for manipulation relative to the super-user designed $o$
	\STATE Save entry \texttt{(k-id,o,l)} in database table \texttt{virtual space-access model} 
	\FOR{each $d\in k_d$}
		\STATE Move selected subtree through each direction $d \in k_d$ from the ``neutral'' pose (see Fig. \ref{same_pose}) to the pose that should be associated with $k_{max}(d,o)$ where pose is given by $(p_1,...,p_M)^T$ and entry $d$ is stored in \texttt{virtual space-access model} table
		\FOR{each pose specified by the super-user}
		\STATE Save each pose in database table \texttt{pose} entry \texttt{(p-id,k-id,p1,...pM)}, entering a \texttt{null} value for any $p\in P$ that is not part of the active $l$; note: the assigned sequential integer corresponding to $s\in k_s$ along with \texttt{k-id}, forms the table's primary key, (\texttt{p-id}, \texttt{k-id}). The primary key of \texttt{vsam} is the foreign key for \texttt{pose}, \texttt{k-id}
		\ENDFOR
		\STATE The last pose for each is saved as $k_{max}(d,o)$
	\ENDFOR
\ENDFOR

\end{algorithmic}
\begin{algorithmic}[1]
\ENSURE 
\STATE User specifies high-level spatial command $(o,l,d,s)$ via a particular interface.
\vspace{-.05in}\begin{equation}p_{next}=f_{SB}(o,l,d,s)\end{equation}\vspace{-.2in}
\STATE Platform moves from current pose, $p_{current}$, to next pose, $p_{next}$.
\end{algorithmic}
\end{algorithm}

	\begin{figure*}[ht]
		\centering		
		\begin{minipage}[c]{.3\textwidth}
			\centering
			\subfloat[\label{baxter}]{\includegraphics[width = 1\textwidth]{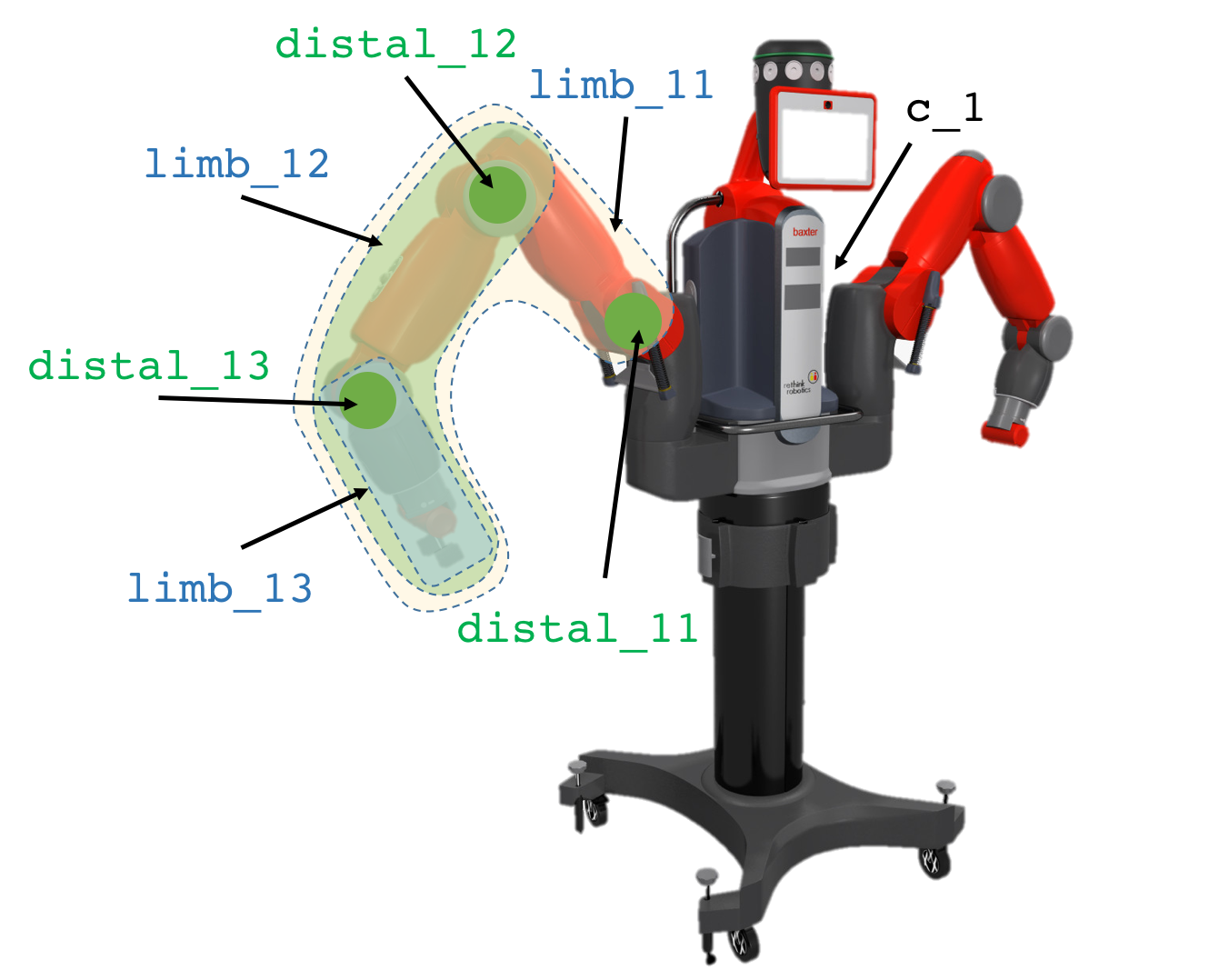}}
		\end{minipage}
		\begin{minipage}[c]{.3\textwidth}
			\centering
			\subfloat[\label{NAO}]{\includegraphics[width=1\textwidth]{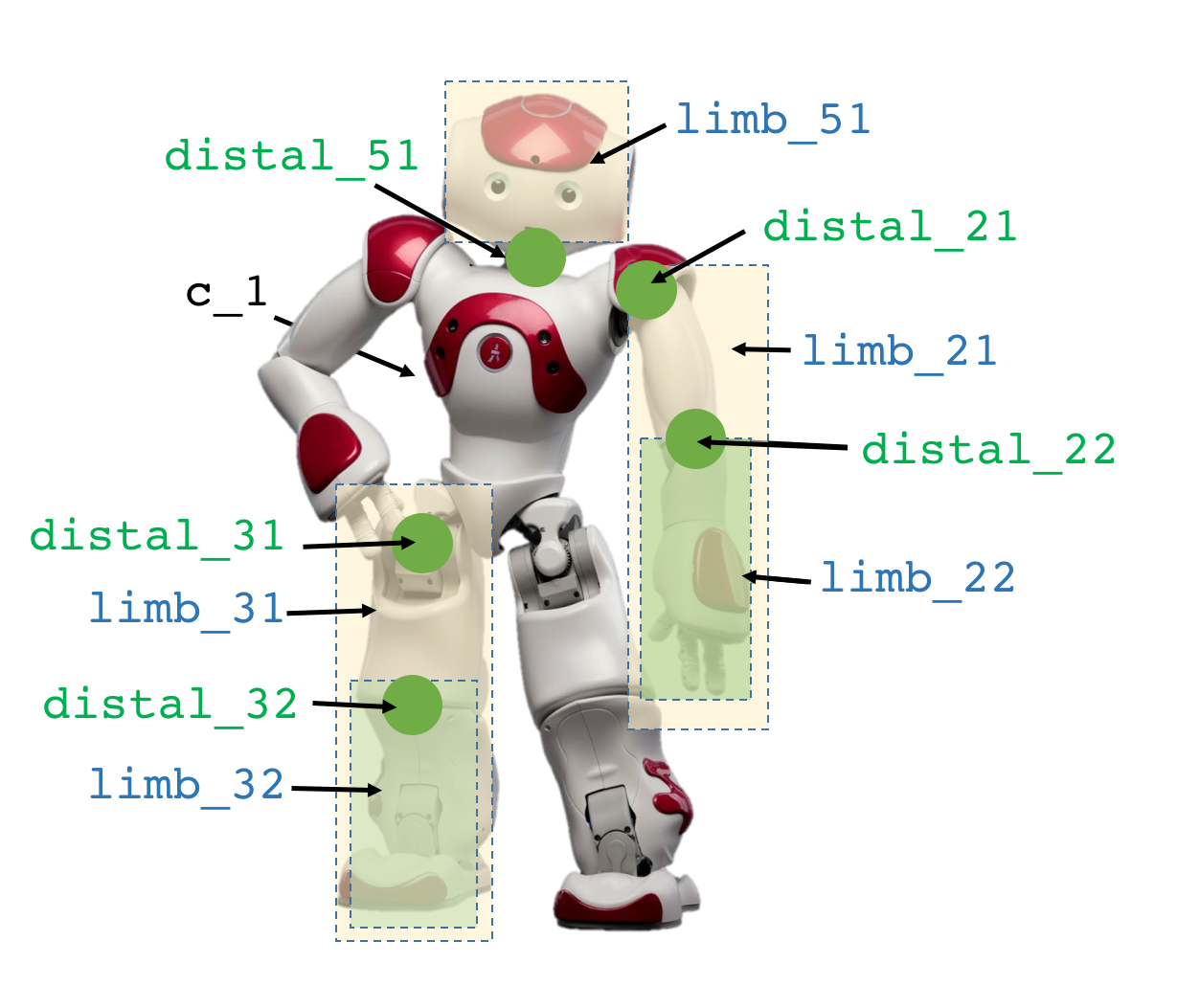}}
		\end{minipage}
		\begin{minipage}[c]{.3\textwidth}
			\centering
			\subfloat[\label{youBot}]{\includegraphics[width=1\textwidth]{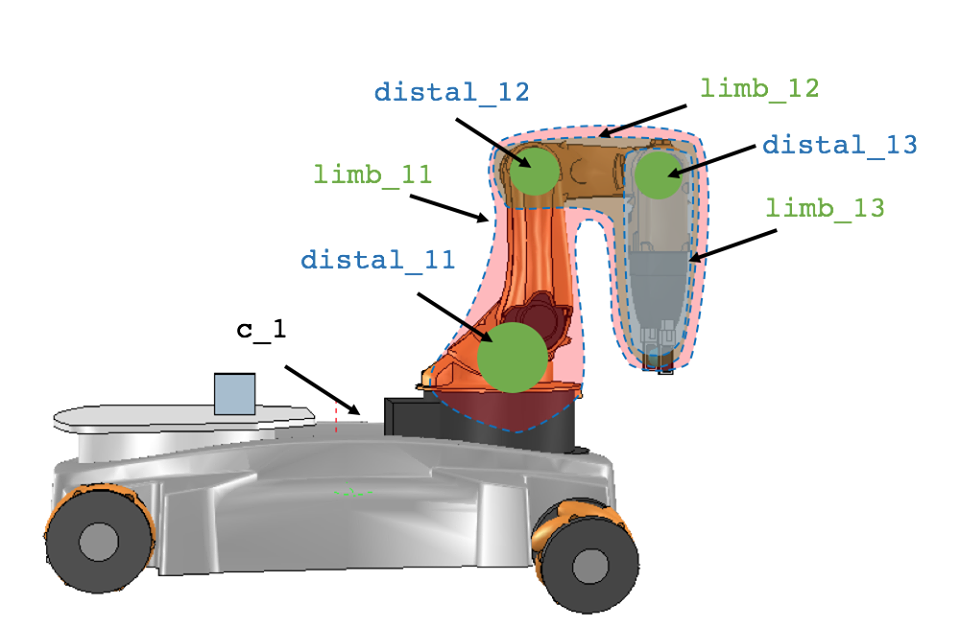}}
		\end{minipage}%

		\caption{ This figure shows the platform labeling scheme for Baxter, NAO, and youBot following the framework detailed in Section \ref{body}. For clarity only half of the part labels are shown here for Baxter and NAO (which have symmetric labels for the opposing limb). Blue text indicates labels for link groups and green text are labels for VSAM origins at joint positions.}	
		\label{body_labels}
		\vspace{-0.15 in}
	\end{figure*}
	
\subsection{Database Design}
The database consists of two entities (to be implemented as two tables) \texttt{vsam} and \texttt{pose} as shown in Fig. \ref{library} and will be implemented so as to avoid redundancy in the final database architecture as in \cite{ullman1982first}.  These entities have a many-to-one relationship where many poses get associated to a single origin and direction pair and are linked via the \texttt{k-id} and \texttt{p-id} attributes, which correspond to the notion of VSAM size $s\in k_s$.  In particular, \texttt{k-id} is the primary key in the \texttt{vsam} table and a foreign key, enacting the many-to-one relationship, in the \texttt{pose} table.  The \texttt{pose} table has two attributes forming its primary key, \texttt{k-id} and \texttt{p-id}.  In each row, joint positions that are not involved in the movement of the body part are considered null.  Rows can be joined in order to create compound poses involving different body parts around multiple origins.

	
	\vspace{.1in}
\section{Implementation on Three Distinct Platforms} 
\label{implementation}
	
This section will present an implementation on three distinct platforms: Baxter, NAO, and youBot.  While the former two might be considered `humanoids', they have distinct challenges in representing human movement.  The Baxter platform is limited to an `upper body' representation of a human form; the NAO platform is limited in range of motion relative to a human.  Further, extending our previous work to the youBot, provides our method a further challenge: translating human movement to a non-humanoid form. 

	The first step in our framework is to label each platform by creating the sets $(C, J, L, P)$ as explained in Section \ref{body}, creating the eURDF labeling set. Next, super-user parameters, $(k_o, k_d, k_s)$ are detailed for each platform. For spatial directions, $k_d$, Laban's 27 directional-points as shown in Fig. \ref{direction_sym} have been used. These points provide 26 spatial directions relative to the origin of VSAM that a body part will move to generate a pose but are not meant to provide a universal space-access model for all platforms or all tasks.  

	Baxter \cite{robotics2013baxter} is a two-armed, stationary robot designed for repetitive industrial tasks. Baxter is labeled as shown in Fig. \ref{baxter} and is identified by $(C, J, L, P)_{Baxter}$. $C_{Baxter}$ is the set containing all the parts associated with core. Since the core of this platform is fixed, there is only one element in set $C_{Baxter}$. Set $J_{Baxter}$ consists of all six joint locations of Baxter, not including the gripper joints. All labels for body parts are enlisted in set $L_{Baxter}$. Fig. \ref{body_labels} shows overlap in labels for individual links in Baxter's right arm. Set $P_{Baxter}$ specifies a discrete space of possible Baxter poses in terms of 14 joint angles. Thus, we have:
	\vspace{-0.01 in}
\begin{flalign*}
C_{Baxter} &= \{ c\_1\} &\\
J_{Baxter} &= \{distal\_11, \hdots,  distal\_13, distal\_21, \hdots, \\  &\qquad distal\_23 \} &\\
L_{Baxter} &= \{ limb\_11, \hdots, limb\_13, limb\_21, \hdots, limb\_23 \} &\\
P_{Baxter} &= \{p_{1min}, \hdots, p_{1max} \} \times \{ p_{2min}, \hdots, p_{2max} \} \\ &\qquad \times \hdots  \times \{ p_{14min}, \hdots, p_{14max} \}.
	\vspace{0.3 in}
\end{flalign*}

	NAO \cite{gouaillier2008nao} is a humanoid robot that was created for research and educational purposes.  NAO is labeled as shown in the Fig. \ref{body_labels} and is identified by $(C, J, L, P)_{NAO}$. $C_{NAO}$ is the set containing all parts associated with core. NAO also has no independent core movement so there is only one element in set. Set $J_{NAO}$ consists of all nine joint locations of NAO not including the end effector joints. All labels for body parts are enlisted in set $L_{NAO}$. An overlap is shown in Fig. \ref{NAO} in labels for individual links in NAO's left arm and right leg. Set $P_{NAO}$ specifies a discrete space of possible NAO poses in terms of 26 joint angles. Thus, we have:
\vspace{-0.01 in}
\begin{flalign*}
C_{NAO} &= \{ c\_1\} &&\\
J_{NAO} &= \{distal\_11, \hdots, distal\_51,distal\_12, \hdots, \\  &\qquad distal\_42\} &&\\
L_{NAO} &= \{ limb\_11, \hdots, limb\_51, limb\_12, \hdots, limb\_42 \} &&\\
P_{NAO} &= \{p_{1min}, \hdots, p_{1max} \} \times \{ p_{2min}, \hdots, p_{2max} \}\\ &\qquad  \times \hdots  \times \{ p_{26min}, \hdots, p_{26max} \}.
\end{flalign*}

 	youBot \cite{bischoff2011kuka} is a mobile robot with a five degree-of-freedom manipulator attached at the top.  youBot is labeled as shown in the Fig. \ref{body_labels} and is identified by $(C, J, L, P)_{youBot}$. $C_{youBot}$ is the set containing all parts associated with core including the wheels. This robot also has no shape deformation in the core so there is only one element in set. Set $J_{youBot}$ consists of all three joint locations of the robot and does not include the end-effector. All labels for body parts are enlisted in set $L_{youBot}$. An overlap can be seen in Fig. \ref{youBot} in labels for individual links of the arm. Set $P_{youBot}$ specifies a discrete space of possible youBot poses in terms of five joint angles. Thus, we have:
\vspace{-0.01 in}
\begin{flalign*}
C_{youBot} &= \{ c\_1\} &&\\
J_{youBot} &= \{distal\_11,  distal\_12, distal\_13\} &&\\
L_{youBot} &= \{ limb\_11, limb\_12, limb\_13\} &&\\
P_{youBot} &= \{p_{1min}, \hdots, p_{1max} \} \times \{ p_{2min}, \hdots, p_{2max} \}\\ &\qquad  \times \hdots  \times \{ p_{5min}, \hdots, p_{5max} \}.
\end{flalign*}

		\begin{figure}
	\centering
	\includegraphics[width=\columnwidth]{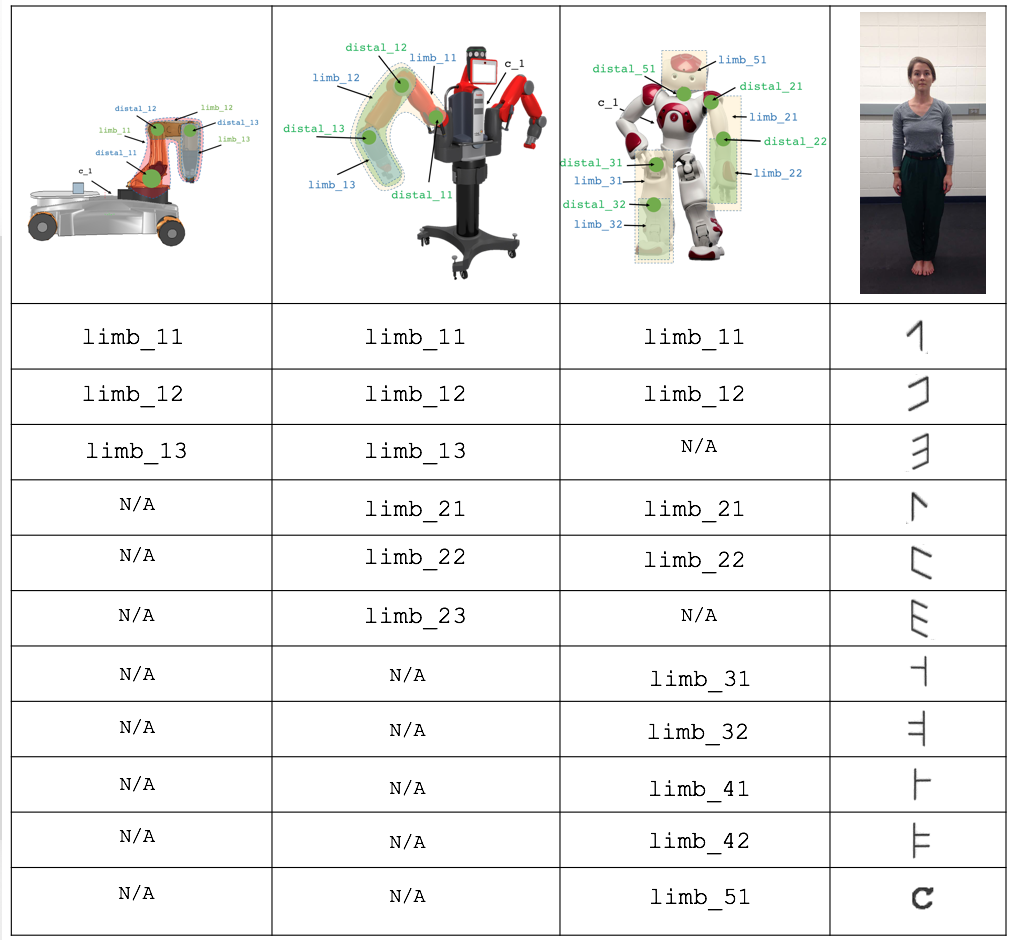}
	\caption{The platform-invariant labeling scheme is relatable to Motif symbols. Just as human body parts are assigned symbols in Motif (without being specific to a particular human), links in a serial link chain can also be assigned labels for a wide range of robots, as shown in case of three here. For example, label \texttt{limb\_12} has meaning in each of these three robots, in addition to being equivalent of right forearm in a human.}
	\label{motif_table}
	\end{figure}

	The super-user specified parameters that define the VSAM for this instantiation are similarly denoted with a subscript: $(k_o, k_d, k_s)_{Laban-26}$. The set of origins for each robot is the same as the set of joints, $J_{platform}$, depicted by circles in Fig. \ref{body_labels}.  Thus, the assignment $k_{o_{Laban-26}}=J_{platform}$ is made for each of the three platforms. 
 Thus, for each Baxter arm, there are three elements in $k_o$ indicated in Fig. \ref{baxter}. For each of NAO limb, there are two elements in $k_o$ indicated in Fig. \ref{NAO}.  Finally, for the youBot there are three elements in $k_o$ indicated in Fig. \ref{youBot}. 
	
Similarly, VSAM sizes are specified in the set $k_{s_{Laban-26}}=\{1, 2, 3\}$, corresponding to near, mid, and far-reach.  Here, we use three sizes, limiting the resolution of the results greatly.  It would be easy to sample with greater resolution, storing the poses in a cloud architecture.
	

	The set of directions used are denoted as $k_{d_{Laban-26}}$ =\{Left-Forward-High, $\hdots$, Right-Back-Low\} which are Laban's 26 spatial directions as described in Fig. \ref{direction_sym}. Note that for $limb\_13$ in Baxter, the most distal link, there are only three spatial pulls namely: Left-High, Forward-Middle, Right-forward-Low. Similarly, for $limb\_23$, there are three spatial directions namely: Right-High, Forward-Middle and Left-Forward-Low. Similarly, for NAO's $limb\_51$ (head), there are only five spatial directions namely: Place High, Place Middle, Place Low, Left-Middle, Right-Middle. Spatial directions that would cause dynamic instability in NAO were not recorded for the lower limbs. Moreover, NAO and youBot are mobile platforms so the method also includes the case for translation in eight directions where $c\_1$ is VSAM's origin as well as \texttt{Body\_part}. These eight directions are the spatial pulls from Middle level shown in Fig. \ref{direction_sym}.
	
\begin{figure}
	\centering
	\includegraphics[width=0.9\columnwidth]{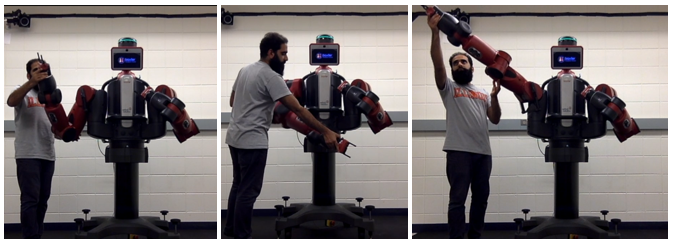}
	\caption[Super-user Manipulating Baxter for Pose Recording]{This figure shows images of a ``super-user'' physically manipulating Baxter's arm to record static poses for its ECL following procedure outlined in Algorithm \ref{alg:main}.}
	\label{ecl-sp}
	\vspace{0.2in}
	\end{figure}

\begin{figure}
	\centering
	\subfloat[][]{\includegraphics[width=1.\columnwidth]{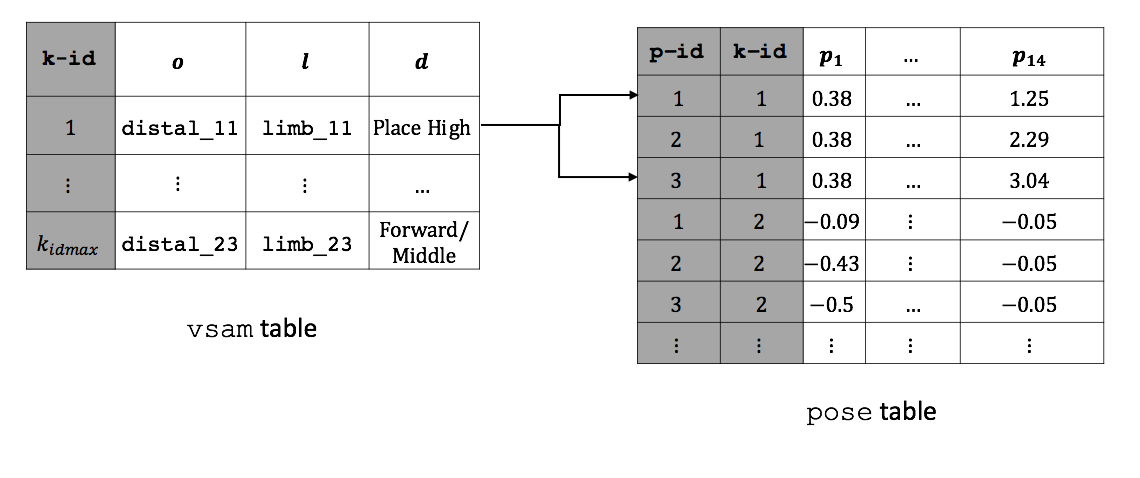}\label{library_baxter}}
	\qquad
	\subfloat[][]{\includegraphics[width=1.\columnwidth]{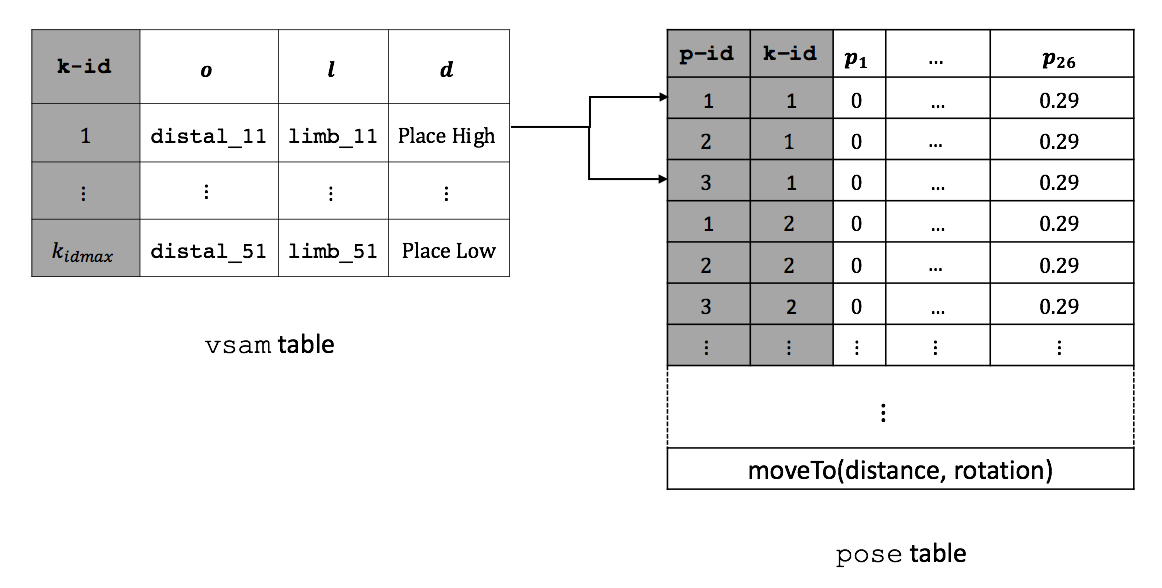}\label{library_nao}}
	\qquad	
	\subfloat[][]{\includegraphics[width=1.\columnwidth]{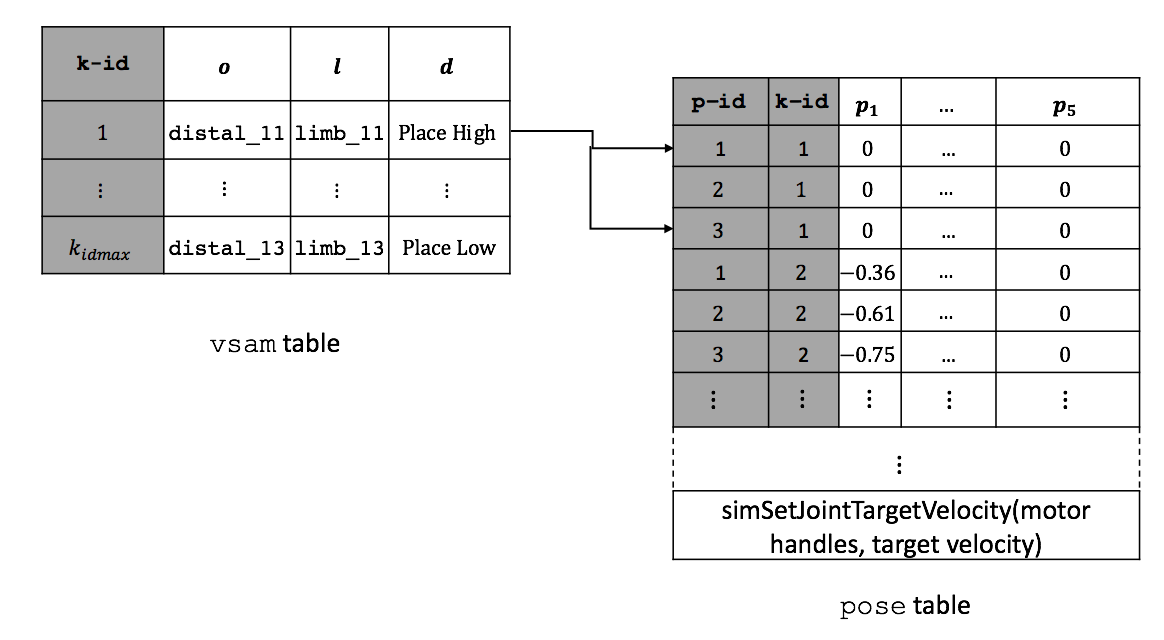}\label{library_youbot}}
	\caption{{This figure show sections of both databases corresponding to the ECL implementation as explained in Fig. \ref{library}: (a) Excerpt from Baxter's implementation. (b) excerpt from NAO's implementation. (c) Excerpt from youBot's implementation.} }
	\label{db}
	\vspace{0.15 in}
	\end{figure}

	Using the labels, $(C, J, L, P)$, of each platform and $(k_o, k_d, k_s)_{Laban-26}$, Algorithm \ref{alg:main} outlines the series of steps performed by the super-user to generate ECL and how a user can generate a configuration pose (for all three platforms) via sparse user parameters, $(l, o, d, s)$. In this case, the ECLs for Baxter and NAO were generated through direct manipulation of the robot limbs as depicted in Fig. \ref{ecl-sp}.  The ECL for youBot was generated through a similar user-driven manipulator process in simulation.  Figure \ref{db} shows excerpt from ECLs of all three platforms. 
	
		\begin{figure}%
		\centering
		\subfloat[][]{\includegraphics[width=\columnwidth]{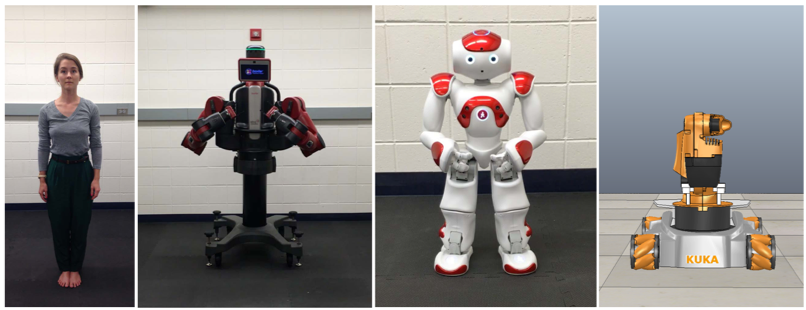}\label{initial}}
		\qquad	
		\subfloat[][]{\includegraphics[width=\columnwidth]{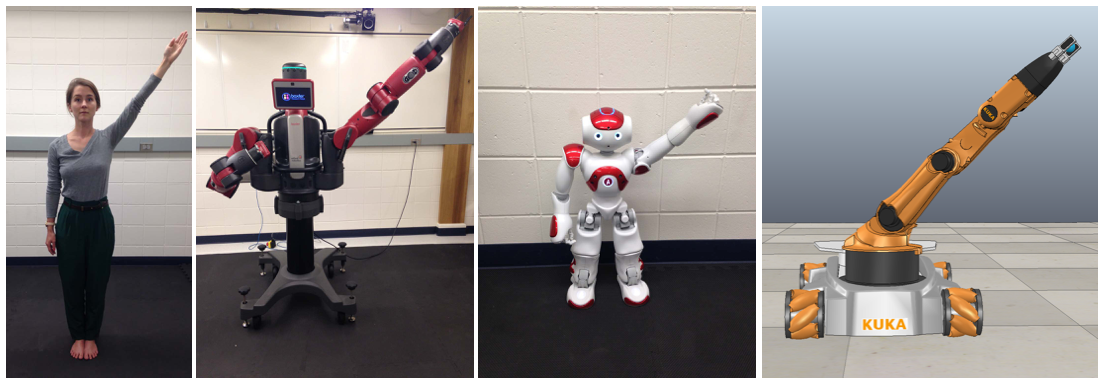}\label{pose}}		
		\caption{Poses defined by architecture: (a) shows the neutral home pose for a human mover, Baxter, NAO and youBot, in this implementation; (b) shows the output pose for the same user input, ($distal$\_11, $limb$\_11, Left-High, 3).  Even though the platforms are physically different, they all raised leftmost, upper subtree of their structure up and to the left direction, to the maximum extent for the given user command. The mapping defined here is extensible and flexible.} %
		\label{same_pose}%
		\vspace{0.15 in}
	\end{figure}

Using ECLs created for NAO, Baxter, and youBot, a user can retrieve any pose by specifying sparse parameters. For example, for a high-level command of raising right arm in Left-High direction, the user specifies the body part labeled $limb\_11$, from an origin at $distal\_11$ with spatial direction Left-High and size of 3 (maximum size stored). The output of specifying these parameters can be observed in Fig. \ref{same_pose} as compared to a human subject executing the same high level instruction. 
A pose of Left-High for body part labeled $limb\_11$ in each of Baxter, NAO and youBot robots is observed. The platform-invariant nature of this framework is highlighted in using the same set of commands to produce a pose that would otherwise require different scaling of parameters for physically different platforms. See Fig. \ref{motif_table} for the symbolic correspondence to these commands.

  \begin{figure}[h!]
	\centering
	\includegraphics[width=\columnwidth]{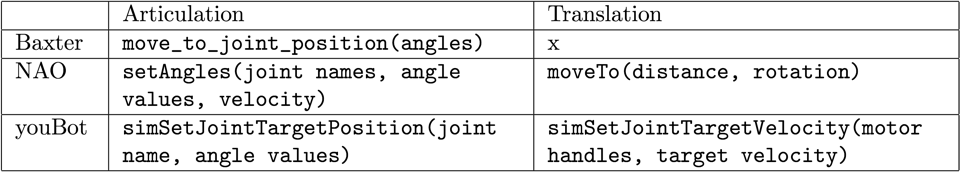}
	\caption{Library commands used to generate motion.} \label{SDKtable}
	\vspace{.1in}
	\end{figure}  

To implement the selected sequence on a robot, we leverage existing software development kits (SDKs).  Both Baxter and NAO have native SDKs and the youBot is simulated in Coppelia Robotics V-REP simulation software. The desired motion (articulation and translation) is executed on platforms using the commands listed in Fig. \ref{SDKtable}

	
	\section{Validation via Lay User Movement Design} \label{userstudy}
	To evaluate the effectiveness of the proposed framework for generating platform-invariant movements, we conducted a questionnaire based user study. Nine participants (7 females and 2 males between ages 19 and 29 years) from the University of Illinois were recruited by advertising through fliers. The participant's departments ranged from psychology to mechanical engineering.  Five out of nine participants answered as having no prior knowledge of LBMS, and out of the remaining four, the most well-versed participant had received less than ten hours of training with a Certified Movement Analyst before. 

	
	
	
	The study was conducted in two sessions. In the first part, an expert in LBMS, a Certified Movement Analyst (CMA), conducted a 30-minute training session with the volunteers to equip them with vocabulary to describe movement in a compact manner. Relevant concepts about Space and Body along with their symbolic representation were explained. Participants moved around in three different reach spaces around their body: near-reach, middle-reach and far-reach. Moreover, Laban's twenty-seven spatial pulls as depicted in Fig. \ref{direction_sym} were introduced as a spatial representation in which a motion takes place. A simplified Motif representation was explained using the symbols that align with our architecture. 
	
During this training, participants practiced moving around in their environment experimenting with different spatial pulls, body parts involved and vicinities in which movement can happen.  {For example, moving in response to a spatial command `Left', a human will naturally translate left, but they will also \textit{reach their arm to the left}.}  This is a shape deformation in response to a spatial command.  This is the natural human behavior that is formalized in LBMS and leveraged for kinesthetically meaningful control architecture here. Therefore, this training accessed much of the kinesthetic intelligence human movers already possess and repackaged it with a formalized language.

	After the training and practice session, the participants were requested to `create a movement phrase' in the context of either a restaurant hostess, tour guide or dancer, incorporating four movements in their sequence.  Then, participants were asked to label this phrase of four movement in terms of the vocabulary they had just learned.  Unbeknownst to the participants, this vocabulary corresponded to robot mappings shown in Fig. \ref{motif_table}. Four participants selected the `tour guide' context, two selected `restaurant hostess' and the remaining three picked `dancer'. 
	
\begin{figure}
	\centering
	\subfloat[][]{\includegraphics[width=0.85\columnwidth]{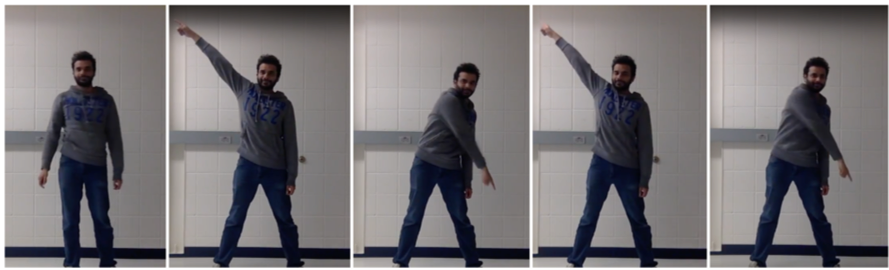}\label{user}}
	\qquad
	\subfloat[][]{\includegraphics[width=.9\columnwidth]{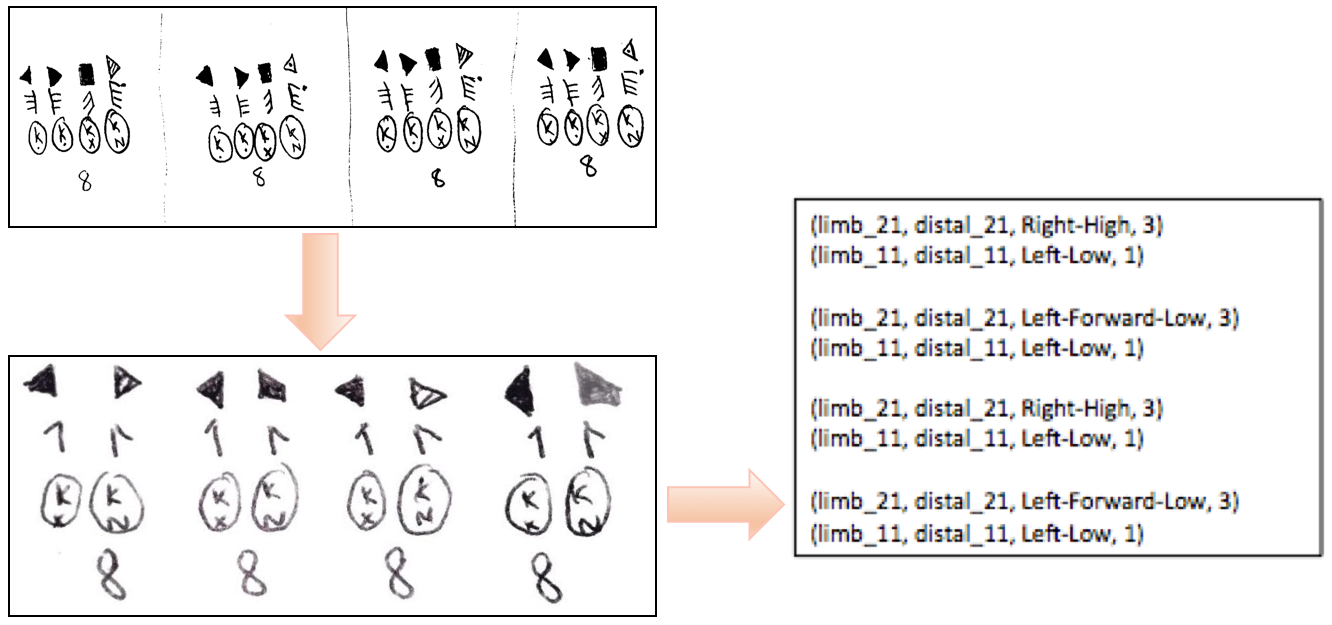}\label{motif}}
	\qquad
	\subfloat[][]{\includegraphics[width=0.85\columnwidth]{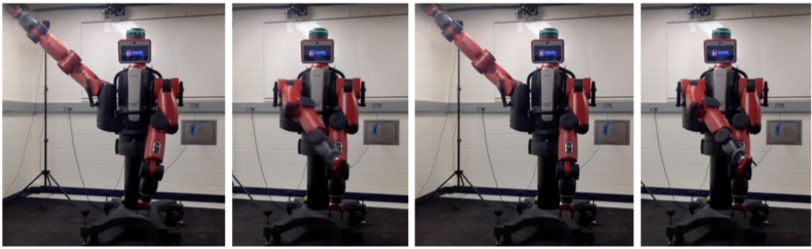}\label{sequence_baxter}}
	\qquad
	\subfloat[][]{\includegraphics[width=0.85\columnwidth]{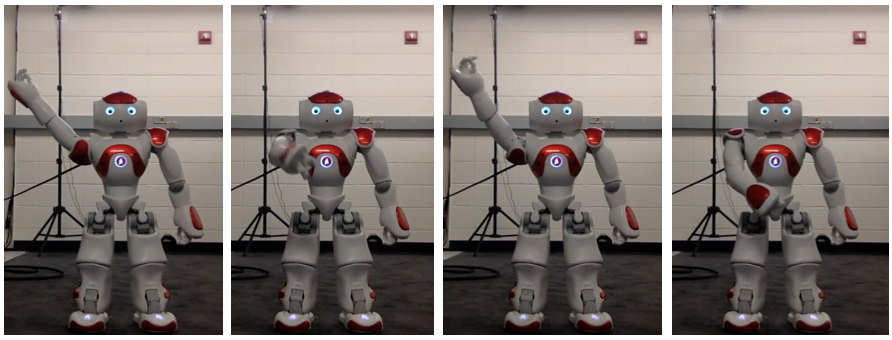}\label{sequence_nao}}
	\qquad
	\subfloat[][]{\includegraphics[width=0.45\columnwidth]{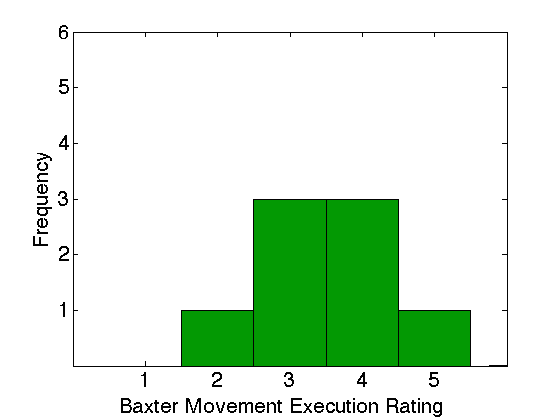}\label{rating_baxter}}
	\subfloat[][]{\includegraphics[width=0.45\columnwidth]{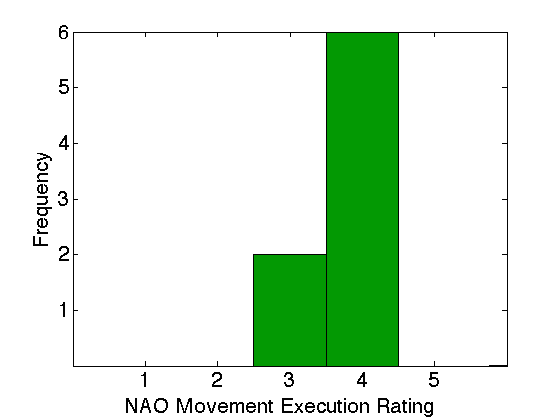}\label{rating_nao}}
	\qquad
	\subfloat[][]{\includegraphics[width=0.45\columnwidth]{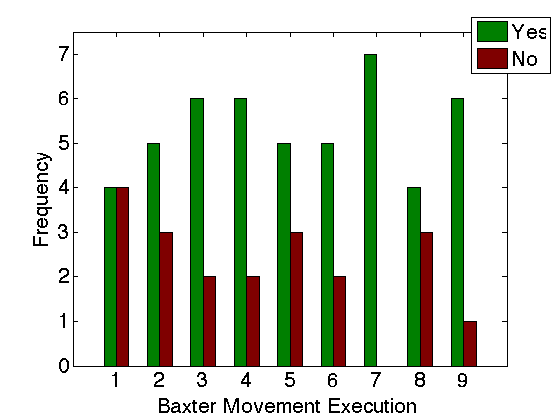}\label{all_baxter}}
	\subfloat[][]{\includegraphics[width=0.45\columnwidth]{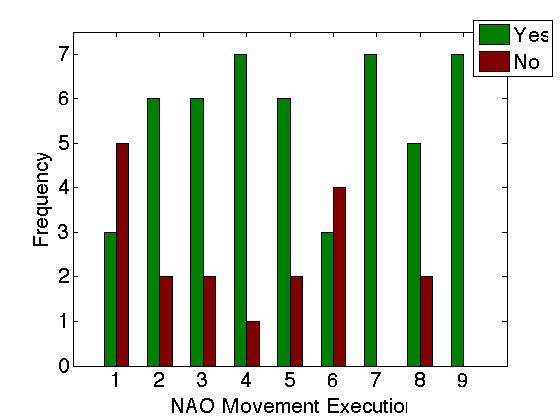}\label{all_nao}}
	\caption{The results from the first user study.  (a) Snapshots of participant's original movement sequence (in the context of `dancer'); this is Movement Sequence 9.  (b) Participant's Motif and corrected Motif.  (c) Snapshots of execution of sequence on Baxter.  (d) Snapshots of execution of sequence on NAO.   (e), (f) All participants' ratings of both platforms execution of their own sequence.  (g), (h) Participants' binary choice of whether the original sequence matched the robot execution across all sequences. 
	}
	\label{userdata}
	\end{figure}

	After designing the movement, the participants had to provide a written description in English, a motif representation and a video recording of their movement sequence. Figure \ref{user} shows images of movement sequence collected from one participant and Fig. \ref{motif} shows their Motif representation. At the end of this session, participants filled in a questionnaire to give feedback on the training. 
	
	Afterwards, the Motifs were corrected by someone with approximately 10 hours of LBMS training (not a CMA expert) if needed, e.g., a symbol for fingers was replaced with one for whole arm when the video conflicted the Motif. After correction, the Motifs were converted to high-level user command of $(l,o,d,s)$. Figure \ref{motif} shows an example of motif provided by the participant and the resulting user commands. 
	 From these input commands the proposed framework generated movement on two of the robotic platforms: NAO and Baxter. Figures \ref{sequence_baxter} and \ref{sequence_nao} shows series of images captured from movement execution on the two platforms for user commands in Fig. \ref{motif}. 
	 
	
	Next, for the second part of the user study,  participants were shown videos of movement execution on both robots and were asked to evaluate if the movement looked the `same' as what they intended. Each participant not only viewed his/her own movement but the movement sequences of other participants. This part of the user study was conducted remotely using SurveyMonkey questionnaires. The questions asked for each robot's execution were:
	
	 \begin{itemize}
 	 \item Did the platform execution capture your intended style of movement? In other words, would this movement sequence look appropriate in your selected context? \textit{[Yes or No]}
 	 \item What important aspect did the execution miss?  What are the similarities between your execution and platform's execution? \textit{[Free response]}
 	 \item Remembering that the robotic platform is limited in its movement capability, please rate how similar the execution seemed `in spirit' to your movement in the video. \textit{[1 (Not Similar), 2, 3, 4, 5 (Very Similar)]}
	 \end{itemize}
		The average rating for participant satisfaction about the training was 4.6 out of 5 with a standard deviation of 0.7 indicating that participants felt confident to create movement sequence for given context.  This minimal training could easily have been extended to a few hours, representing a very reasonable time frame for training an operator on a new system.  Given the limited training session, users did make some errors in their usage of the taxonomy, which were corrected by the team via video observation.
		

	
	
	 Eight out of the original nine participants that designed the movement sequence participated. All eight of them responded `Yes' that the execution captured their intended style of movement for NAO's execution. However, for Baxter, two participants out of the eight responded `No'. Participants justified this rating by writing that the execution was missing `curv[ature] of the arms' and `movement of head'.
	 
	   The common comments about similarities between the movements were: `Reach of the arms and closeness', `same limb and head movements (for NAO)', `the poses taken by the robot were very similar' and `the poses looked like dance moves'. On the other hand, some of the comments about the differences in execution were: `speed: robot is too slow, hands: robot does not look like it is grasping objects', and `there's no legs or head movement'. 
	 
	 The results of participants rating similarity between their own movement and robot's movement execution is captured in Figs. \ref{rating_baxter} and \ref{rating_nao}. For Baxter, the highest and lowest rating received was 2 and 5 respectively. For NAO's execution, six out of the eight participants gave the executions a rating of 4 with the remaining giving a rating of 3. One reason from 75\% of participants gave NAO a high rating of 4 could be that it is perceived as more similar in shape to a human than Baxter.
	 

			 	\begin{figure*}[ht]
	\centering
	\subfloat[][]{\includegraphics[width=\textwidth]{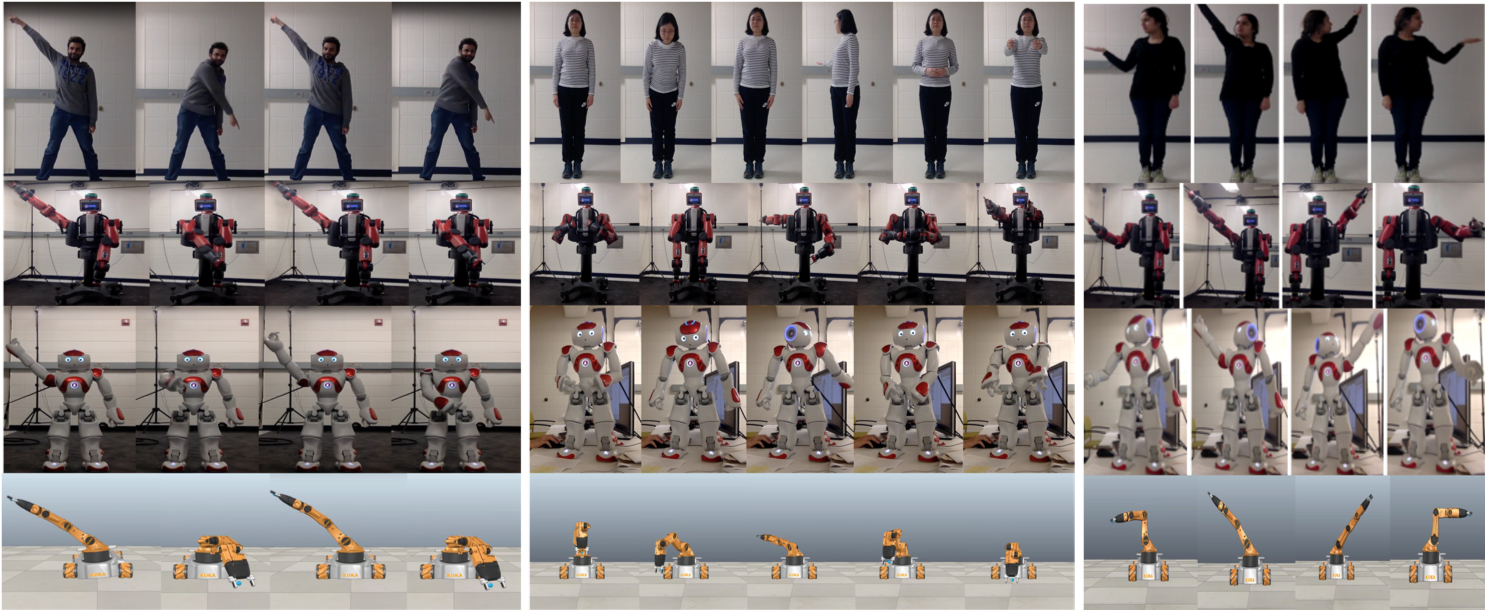}\label{ms_967}}
	\qquad
		\subfloat[][]{\includegraphics[width=\textwidth]{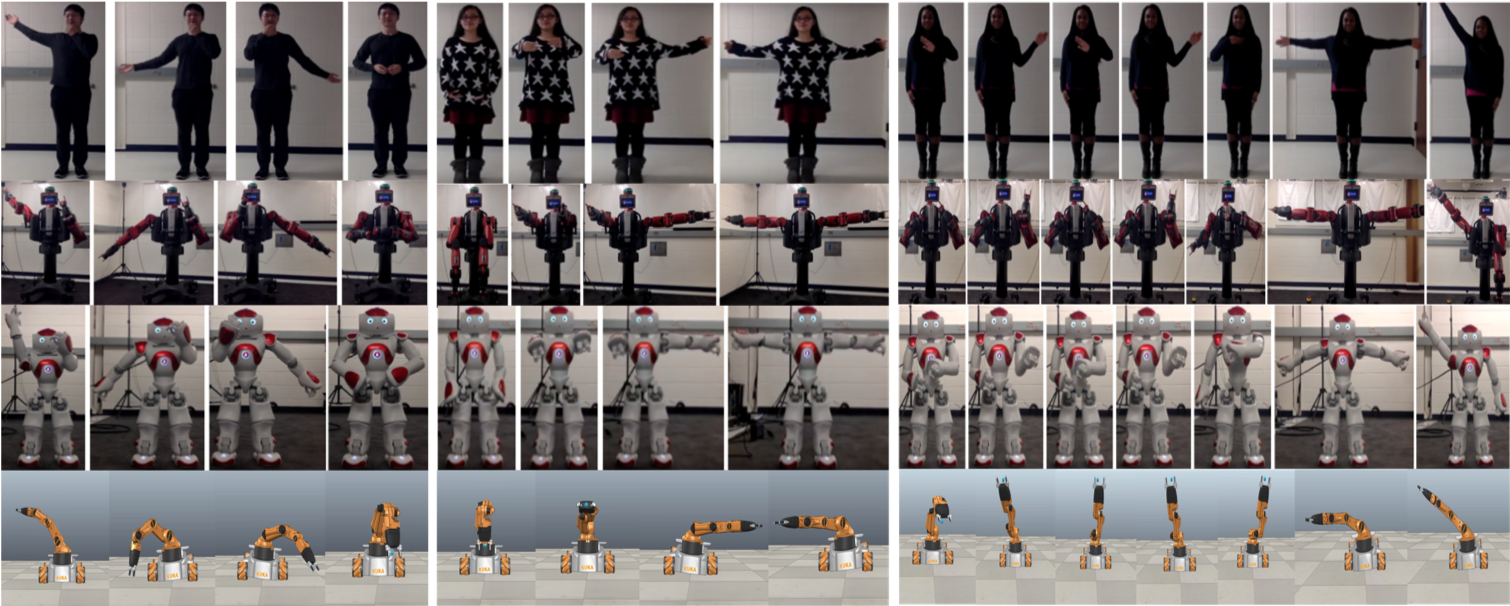}\label{ms_254}}
	\caption{Six examples of the framework.  (a) Left to right: Movement Sequence 9, 6, and 7.  (b) Left to right: Movement Sequence 2, 5, and 4.   Full details of the data collected can be found in \cite{sher2017embodied}. 
	}
	\label{userdata2}
	\end{figure*}

	The participants also evaluated other participant's movement execution on the robots. Figures \ref{all_baxter} and \ref{all_nao} show histogram of how many participant answered `Yes' or `No' if the robot execution was successful in capturing the style of human actor's movement sequence. For both platforms, Movement Sequence 1 received more `No' than `Yes'. For this execution, there was leg movement involved that was not captured on the platforms. The super-user could have incorporated that in the ECL to improve user response. Sequence 6 in NAO's case also had higher red bar. Participants mentioned that ``handing out the menu (movement) didn't seem to capture the essence of it''. This indicates that functional tasks such as handing out menus are not the strength of an ECL based on Laban's 26 spatial directions. All remaining movement sequences for both platforms had higher green bars than red bars. For Movement Sequence 7 for both platform and Sequence 9 for NAO, all participants unanimously agreed that the execution on the platform captured the intended style of movement of the human actor's execution. 
	
	A majority of the participants agreed that the movement execution on the robots captured the movement intent of the human actor. Specifically, 15 out of 18 executions (83\%) were rated as preserving the original movement phrase, validating that the proposed framework is successful in generating platform-invariant movement via high-level commands.  Further, our users submitted to a relatively short training session and did not need to learn technical details of the architecture in order to participate.
	
%


		\section{Validation via Lay User Observation}\label{userstudy2}
		In the user study described in the previous section, we collected movement data from nine participants. The participants designed a movement sequence consisting of four movements in context of either a dancer, a restaurant hostess or a tour guide. The participants provided movement data as a video recording of movement, a Motif representation as well as a written English description. The user provided Motifs were corrected to match the movement video provided. These Motifs were translated to four user parameters, $(l,o,d,s)$, to generate movement now on three robotic platforms: Baxter, NAO, and youBot.  The youBot was added to test performance on a non-anthropomorphic platform.
		
		In this study, the participants watched videos of movement execution on these three robots and evaluated their performance.  A new group of volunteers of age 18 years and older were recruited from University of Illinois at Urbana-Champaign. 
		The participants were from a wide variation of educational background such as History, Psychology, Dance, Bioengineering, Global Studies, Business Management, Molecular Biology, Mechanical and Electrical Engineering. This represents a mix of technical as well as non-technical backgrounds.  This study was conducted online via SurveyMonkey and it took less than two hours to finish. The participants were compensated \$30 in Starbucks or Amazon gift cards for their time. 
		
		The participants in this user study did not receive any training in the LBMS system, Motif, or movement priming of any kind. Participants were asked their level of expertise of LBMS and 16 out of 18 participants responded that they had no prior knowledge of it. Other movement training experience ranged from training in dance, tennis, soccer, cricket, swimming, Tai Chi, and cross country running. 
		
The study was designed such that all participants were given a brief primer about the robotic platforms under consideration so all participants had basic knowledge about the movement capabilities of these platforms irrespective of their background. The participants were quizzed afterwards  and correct answers were presented to ensure that they understand the capabilities of each platform. Next, the volunteers were showed each movement sequence being performed by a human actor and given the context around which it was created: either `tour guide', `restaurant hostess', or `dancer'. The participants were quizzed about the context of each movement to ensure that they had noted the context of the movement. 
		
		Then, the participants watched three different platform's movement execution corresponding to the human actor's movement and answered the following questions:
		\begin{itemize}
			\item Did the platform execution capture the intended style of movement of the human actor? In other words, would this movement sequence look appropriate in the given context? \textit{[Yes or No]}
			\item What important aspect did the execution miss?  What are the similarities between human actor's execution and platform's execution? \textit{[Free response]}
			\item Remembering that the robotic platform is limited in its movement capability, please rate how similar the execution seemed `in spirit' to human actor's movement in the video. \textit{[1 (Not Similar), 2, 3, 4, 5 (Very Similar)]}
		\end{itemize}
		
\begin{figure*}[ht]
			\centering		
			\begin{minipage}[c]{.3\textwidth}
				\centering
				\subfloat[\label{baxter2}]{\includegraphics[width = 1\textwidth]{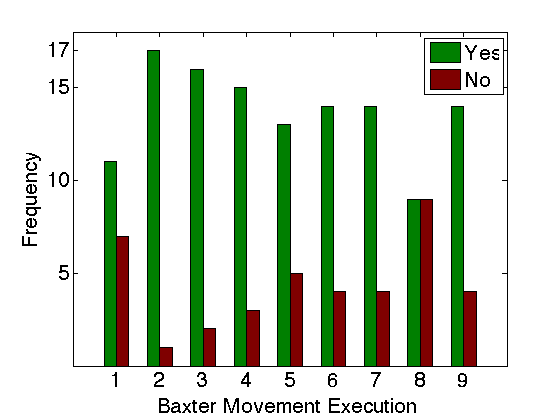}}
			\end{minipage}
			\begin{minipage}[c]{.3\textwidth}
				\centering
				\subfloat[\label{nao2}]{\includegraphics[width=1\textwidth]{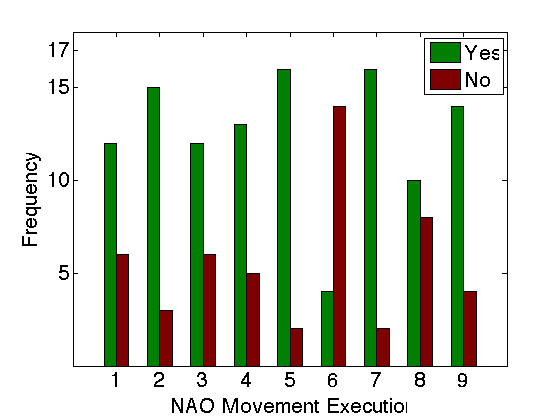}}
			\end{minipage}
			\begin{minipage}[c]{.3\textwidth}
				\centering
				\subfloat[\label{youbot2}]{\includegraphics[width=1\textwidth]{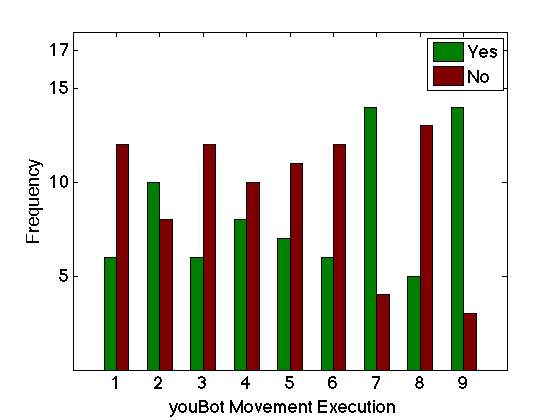}}
			\end{minipage}%
			
			\caption{ This figure shows histogram of binary response of user if the robotic platform movement was similar to human actor's movement. For Baxter, in all of the movement sequences more participants responded ``Yes'' than ``No''. For NAO, all except one movement sequence had more ``No'' than ``Yes''. However, for youBot, only for three out of nine movements, more participants responded ``Yes'' than ``No''.  }
			\label{userstudy2-binary}
		\end{figure*}

\begin{figure*}
			\centering		
			\begin{minipage}[c]{.3\textwidth}
				\centering
				\subfloat[\label{baxter2rate}]{\includegraphics[width = 1\textwidth]{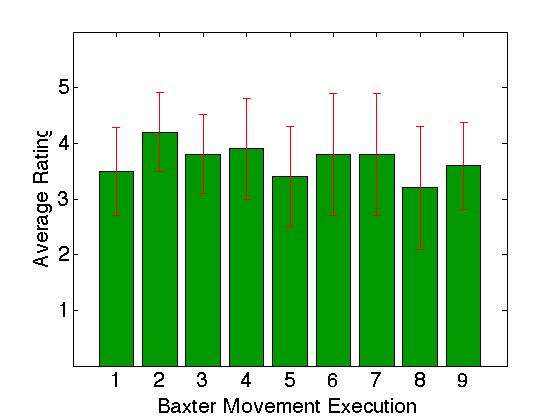}}
			\end{minipage}
			\begin{minipage}[c]{.3\textwidth}
				\centering
				\subfloat[\label{nao2rate}]{\includegraphics[width=1\textwidth]{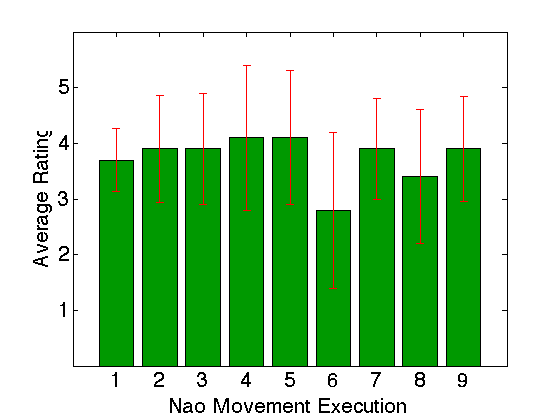}}
			\end{minipage}
			\begin{minipage}[c]{.3\textwidth}
				\centering
				\subfloat[\label{youbot2rate}]{\includegraphics[width=1\textwidth]{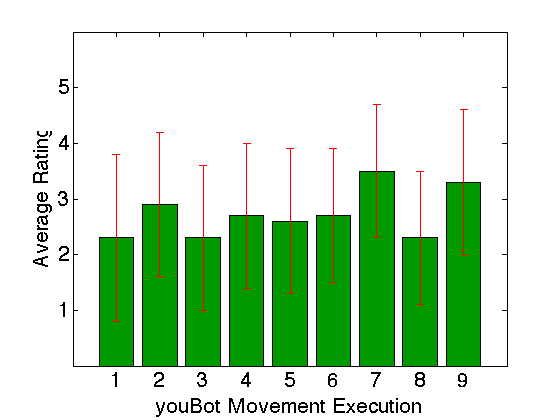}}
			\end{minipage}%
			
			\caption{ This figure shows a bar chart of average rating for each platform. The rating is on a scale of 1 to 5 where 5 is the highest rating that can be awarded. For Baxter and NAO, all movement sequences received a rating of 3 or higher matching the participants response in Figure \ref{userstudy2-binary} whereas for youBot only movements 7 and 9 received a high rating of 3 and above. These two movement sequences were made in context of dance. }
			\label{userstudy2-avgrating}
		\end{figure*}

		
		As shown in Figure \ref{baxter2}, for all movement sequences for Baxter, more participants responded ``Yes'' to the binary question on whether the platform captured the intended style of movement. For Movement Sequence 2 in Baxter, all participants responded ``Yes'' to the question. This sequence was in context of a tour guide and was composed of directional gestures, which is one of the strengths of this particular ECL. Figure \ref{userstudy2-avgrating} shows average rating of similarity between the human actor's execution and the platform's execution for all movement sequences. The response is given on a scale of 1 to 5 where 5 is highest rating given for movement sequence that is ``Very Similar''. Similar to the trend seen in Figure \ref{userstudy2-binary}, all movement sequences for Baxter received high average rating of 3 and above. 
		
		The participants' comments provided more insight into the similarities and differences that they observed. The participants noted that Baxter was successful in executing the correct direction of the arm movements of the human actor with comments such as ``the arms went in the correct directions'', ``Baxter was able to hit all of the angles like the actor'', ``it got the exact poses'', ``Everything went in the same directions so it looked like a dancer'' and so on. 
		
		On the other hand, the differences that the participants noted had a common theme of movements that were not physically feasible for the platform. The participants observed ``obviously Baxter missed the all of the leg motions'', ``Without the torso movement, the movement seemed more limited'', ``Baxter is unable to mimic leg or torso motion'', etc. The participants were also not satisfied with the speed of execution for the robot and mentioned ``(the platform) did not move as quickly as the dancer'' and ``The speed of arm is very slow''. We choose not to modify the speed of the videos to provide participants correct depiction of how the robot moves in real life. Overall, participants seemed satisfied with the arm movement of Baxter.
		
		For NAO, except for Movement Sequence 6, more participants responded ``Yes'' than ``No'' to the same question for this platform as observed in Figure \ref{nao2}. In Sequence 6, NAO rotates its arm in an undesirable manner compared to how humans rotate their arms that could be one of the reasons for this response. Similar to Baxter, NAO also received a high average rating of 3 and above for all the movement sequences as shown in  Figure \ref{nao2rate}.
		
		Similar to Baxter, for NAO participants agreed that the arm and head movements were translated accurately. Some comments were ``The arm positions were all very close to exactly where the human actor placed her arms'', and ``extension of arm and bending them back (are similar to human actor's movement)''. For the differences the participant commented of missing leg movement. Overall, participants identified the limitation of movement for the robot compared to human movement and commented ``Very similar, considering the robot's limitations''.
		
		On the other hand, for youBot which is is not a humanoid robot, for only three movements out of nine, more participants responded ``Yes'' than ``No'' for the binary response question as shown in Figure \ref{youbot2}. There could be several reasons for this response from the participants. Since all movements used both arms, the arm movement couldn't be mapped directly to the robot. Therefore, in the creation of the ECL, we made arbitrary choice of selecting the human arm that mapped to the robot's arm.  As seen in  Fig. \ref{youbot2rate}, for youBot, only two movement sequences made in context of dance and tour guide received a high rating of 3 and above. These two sequences comprised of only one arm movement at a time making that arm the obvious choice for the super-user to translate to the youBot's single arm. It is also interesting to note that average rating for all movement sequences for youBot is greater than 2 implying that even for the worst case the user observed some similarities between the movement of the non-humanoid platform and human actor's movement.
		
		In open-ended comments on similarities participants agreed that given the capabilities of the robot it performed movement in the correct direction. The participants commented ``Given its limitations, it did not miss anything'', ``youBot's arm went in the same general direction of the human actor's right arm'', ``The arm is in the right angles'' and so on. The participants also pointed out that they struggled in seeing similarities in movements where the actor was moving both hands at the same time (which of course cannot be implemented on a platform with a single appendage). Some of the general comments about those movements were ``hard to compare with one arm and no body reference  can't do both arms front'', ``still hard to compare because only one arm'' and ``the lack of two limbs makes it very difficult to emulate the human motion properly''.
		
		Another interesting comment that was observed for youBot was that for movements involving human actor kneeling, the participants noted that it could be observed for the robot as well. The comments were ``youBot seemed to attempt to crouch down to mimic the human actor kneeling'', and ``youBot achieved the level changes of the human actor''. It is interesting because youBot is not capable of changing levels however the users perceived the change of level because of the movement of the arm. This was not perceived for Baxter or NAO movement executions.  We believe this may be because more ``risky'' movements, causing the youBot to tip momentarily, were instantiated for this platform (which was only operating in simulation), causing the perceived effort by the platform to be higher.
		
		Interestingly, the participant with the most dance training (17 years as entered in their background) preferred the movement of youBot over the Baxter platform, favored by the majority of the participants.  This participant noted the difference in timing and their perceived level of exertion of the overall platform.  For one of the movement sequences with a `tour guide' context, a participant wrote: ``Assuming the actions are for pointing, the youBot does this very well by drawing attention to things. Its full attention appears to be going to something, asking you to do the same. This might even be more effective than the actor's execution because the attention inherently belongs to the tour guide and not the object of attention.''  Similarly, this could be due to the fact that the edges of performance of the platform were pushed in simulation.
		
		Overall, 89\% of movement sequences received a higher number of ``Yes''s than ``No''s for movement similarity for Baxter and NAO across a broad pool of participants.  Further, even though youBot is not a humanoid robot,  33\% of movement sequences received more ``Yes'' than ``No''. The participants' qualitative comments provided insight into the various factors that contributed to the low score such as the type of movement being executed.

		\section{Comparison to Other Methods} \label{comparison}
		
		Our work here is not meant to replace common tools in robotics like inverse kinematics (IK) and motion planning for manipulators.  Indeed our work \textit{requires} this prior work to function (as is evident in Fig. \ref{SDKtable}).  By wrapping a high-level framework around this prior work in robotics, we have demonstrated a framework that is scalable, extensible, and usable.  Other high-level frameworks were introduced in Section \ref{introduction}, and we will compare our work to two such frameworks in detail here.
		
\begin{figure}[h!]
	\centering
	\includegraphics[width=\columnwidth]{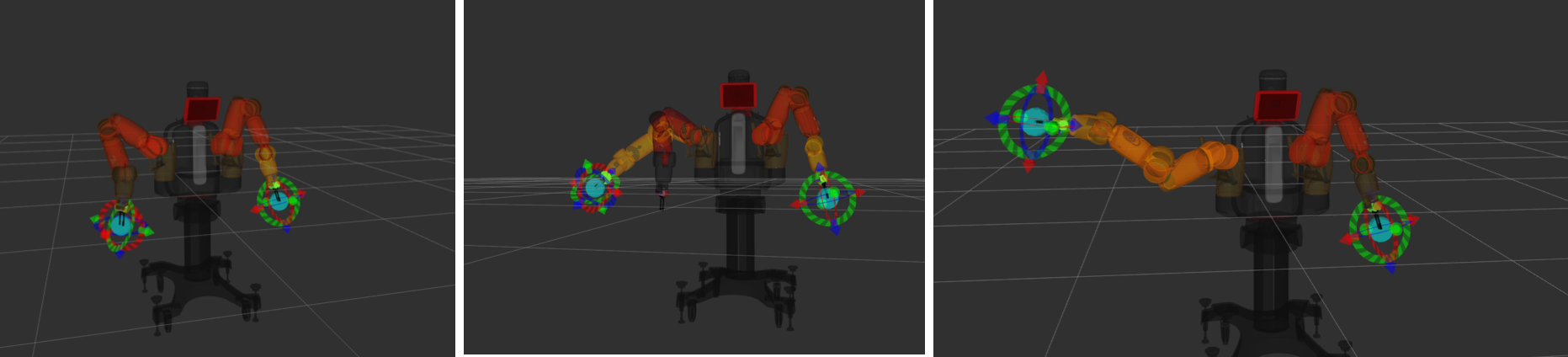}
	\caption{Snapshots from \textit{MoveIt!}, a ROS-based motion planning software.  Arrows stemming from the end effector provide user control inside a simulated environment.} \label{moveit}
	\end{figure}

		First, consider \textit{MoveIt!}, the popular tool inside ROS.  This tool offers a graphical environment in which to develop robot behaviors.  As shown in Fig. \ref{moveit}, users can interact with the endpoints of Baxter's manipulators in this environment.  This environment is visually intuitive and easy to begin with minimal training.  For some behaviors, just clicking on one of the six directional arrows pointing out of the end effector and dragging to a desired location will result in a successful movement.  However, for the behavior shown in Fig. \ref{moveit}, more complex planning must be done by the user in order to accommodate robot actuator arrangements to achieve a desired end pose.  	Here, as with traditional IK planning, users are giving the robot commands in literal, low-level space. So, to achieve this motion, the user must issue a translate command followed by a rotate command followed by another translate command.  Further, this environment does not easily relate to button-based controllers like those used in field environments as discussed for the PackBot in Section \ref{introduction}.
	
		The closest prior work to ours is \cite{salaris2017robot} where Labanotation is used to generate robotic motion for the Romeo movement.  Note that Labanotation is a distinct notation system from Motif, utilizing an anatomical staff to organize the same spatial symbols introduced here.  Moreover, it is a much more detail-oriented system than the Motif shorthand used here.
		The work in \cite{salaris2017robot} focuses on the anatomical staff of Labanotation as well as the spatial symbols from the framework (while other symbols are not considered).  Thus, there is a lot of overlap in our approach as both works focus on the Body and Space categories of LBMS.  One difference is that \cite{salaris2017robot} do not consider kinesphere size as we do here.  Further, dances notated by \textit{expert} Labanotators, with years of experience, on humans are translated to a \textit{specific} humanoid.  Here, we have focused on translation to many platforms after only hours of training on the concepts and symbology at play.  We also include the use of lay viewers to validate that our method has recreated the `same' movement across multiple platforms.  In \cite{salaris2017robot} evaluation of the differences in results is done through expert Labanotation of resulting movements and key differences in execution are highlighted.

	\section{Toward Embodied Platform-Invariant Motion Primitives} \label{conclusion}
	In this paper, we showed how LBMS parameters of Body and Space provide a way of thinking about movement that is inherently ``high-level'' and that has a direct analogy to human movement from a kinesthetic perspective.  In particular, through close study of spatial training in dance and movement studies, we can see how directions in space not only imply a \textit{translation} of a moving platform but also a body-level \textit{articulation}.  
	
	  Thus, our framework provides a unifying command structure for simultaneous articulation and translation through space.  Doing so reduces \textit{command complexity} for controlling platforms that are capable of articulation as well as translation in space such as the iRobot PackBot where currently, a user has to pick one mode at a time. 
	Moreover, in our work, we have differentiated between two groups of users for this framework: a super-user, who would exercise design choices in building the VSAMs and ECLs, and a user, who benefits from this design through the use of a high-level command architecture. 
	
We presented a hierarchical, overlapping labeling scheme that identifies segments of a platform that can be articulated in space.  This scheme is then leveraged to produce a data-driven mapping between high-level spatial commands and platform configuration.  The framework is extensible and can support any future platforms even with increasing (or decreasing) degrees of freedom through super-user design.
	
	Finally, we have demonstrated the efficacy of our method through an implementation for Laban-inspired spatial pulls on the Baxter, NAO, and youBot platforms. The implemented example shows how to produce broad, coarse movements using few parameters. Future implementations could use other specific application-based tasks that requires sequencing of more precise, granular movements using high-level commands. Computational storage is increasingly cheap and accessible (via the cloud); hence the size of the ECL need not be limited. This approach is a step toward reducing kinesthetic mismatch in various human-robot interaction applications and can pave the way for a nuanced, customizable motion imitation scheme.
	
	The user study conducted highlighted the strengths and limitations of the proposed framework.   Nontechnical users felt their designs were implemented successfully on two distinct platforms with minimal training.  However, poses that can be generated are limited to the content of the ECL. This means that a mismatch in notions of space-access between users and super-users gives poor results.  Further, hybrid measures for precise end-effector positioning will likely be needed to extend to manipulation tasks.  
	
	Future work will incorporate further high-level movement ideas into human-operated systems.  For example, the Shape and Effort categories can be used to delineate further information as in \cite{laviers2012ACC} where Effort parameters specify the trajectory of the motion primitive over time.  Further user studies can evaluate the efficacy of using this command architecture in time-sensitive tasks. Virtual reality will help with testing a wide variety of platforms as well as platforms which are currently not physically feasible.
	
	  Hybrid approaches could marry the benefits of joint-angle-by-joint-angle architectures with this high-level approach and could leverage layers of autonomy to further refine the low-level behavior of the platform given a high-level user command. Another way to incorporate this can be by layering it with other motion planning algorithms like those in \cite{7759625}. A possible application in that case is the stylized movement plan for majority of the movement and then shifting to this standard algorithm for finer, more precise movements.  
	  Finally, data-driven learning approaches could be used to automatically determine the exact (or provide a probabilistic definition for the) pose entries in the ECL database.
	  
The overall goal of this work is to provide a flexible system that accommodates the level of control the user wants to exercise over the system by introducing abstraction layers.  As robotic systems become more complex, creating models that facilitate tight kinesthetic intuition between a user and an articulated machine is likely to be something that evolves as broadly as computer interaction paradigms, which have moved from mouse and keyboard to intuitive gestures, over time.  This is necessary to creating robotic systems that accurately capture human intent -- even for nontechnical users -- in a variety of contexts.

	\subsection*{Acknowledgment}
	{This work was funded by DARPA award  \#D16AP00001 and was conducted under UIUC IRB protocol \#16225.}
	
	\bibliographystyle{IEEEtran}
	\bibliography{papers}
\end{document}